\documentclass[10pt,journal,compsoc]{IEEEtran}

\ifCLASSOPTIONcompsoc
  \usepackage[nocompress]{cite}
\else
  \usepackage{cite}
\fi

\usepackage{amsmath}
\usepackage{algorithmic}
\usepackage{array}
\usepackage{url}
\usepackage{times}
\usepackage{epsfig}
\usepackage{graphicx}
\usepackage{amsmath}
\usepackage{bm}
\usepackage{amssymb}
\usepackage{breqn}
\usepackage{subcaption}
\usepackage{verbatim}
\usepackage{algorithm2e}
\usepackage{multirow}
\usepackage{array}
\usepackage{color}
\usepackage{wrapfig}
\usepackage{animate} 
\usepackage[%
    colorlinks=true,
    pdfborder={0 0 0},
    linkcolor=red
]{hyperref}

\newcommand{\boldstart}[1]{\noindent\textbf{#1}}
\newcolumntype{P}[1]{>{\centering\arraybackslash}p{#1}}
\newcommand{\argmin}{\arg\!\min}

\begin{document}
\title{Self-supervised Multi-view Person Association and Its Applications}

\author{Minh~Vo, Ersin~Yumer, Kalyan~Sunkavalli, Sunil~Hadap, Yaser~Sheikh and Srinivasa~G.~Narasimhan
\IEEEcompsocitemizethanks{\IEEEcompsocthanksitem Minh Vo, Yaser Sheikh, and Srinivasa G. Narasimhan are with The Robotics Institute, Carnegie Mellon University, PA, USA, 15213.\protect\\
E-mail:\{mpvo, srinivas, yaser\}@cs.cmu.edu}
\IEEEcompsocitemizethanks{\IEEEcompsocthanksitem Ersin Yumer is with Uber ATG, CA, USA, 94103.\protect\\ E-mail: yumer@uber.com}
\IEEEcompsocitemizethanks{\IEEEcompsocthanksitem Kalyan Sunkavalli is with Adobe Research, CA, USA, 95110.\protect\\ E-mail: sunkaval@adobe.com}
\IEEEcompsocitemizethanks{\IEEEcompsocthanksitem Sunil Hadp is with Amazon Lab 126, CA, USA, 94085.\protect\\ E-mail: hadap@acm.org}
}

\markboth{TRANSACTION ON PATTERN ANALYSIS AND MACHINE INTELLIGENCE, VOL. XXX, NO. XXX, XXX XXX}%
{Minh~Vo \MakeLowercase{\textit{et al.}}: Bare Demo of IEEEtran.cls for Computer Society Journals}

\IEEEtitleabstractindextext{%
\begin{abstract}
Reliable markerless motion tracking of people participating in a complex group activity from multiple moving cameras is challenging due to frequent occlusions, strong viewpoint and appearance variations, and asynchronous video streams. To solve this problem, reliable association of the same person across distant viewpoints and temporal instances is essential. We present a self-supervised framework to adapt a generic person appearance descriptor to the unlabeled videos by exploiting motion tracking, mutual exclusion constraints, and multi-view geometry. The adapted discriminative descriptor is used in a tracking-by-clustering formulation. We validate the effectiveness of our descriptor learning on WILDTRACK~\cite{ETHZ} and three new complex social scenes captured by multiple cameras with up to 60 people ``in the wild''. We report significant improvement in association accuracy (up to 18\%) and stable and coherent 3D human skeleton tracking (5 to 10 times) over the baseline. Using the reconstructed 3D skeletons, we cut the input videos into a multi-angle video where the image of a specified person is shown from the best visible front-facing camera. Our algorithm detects inter-human occlusion to determine the camera switching moment while still maintaining the flow of the action well. 
\textbf{Website}: \url{http://www.cs.cmu.edu/\~ILIM/projects/IM/Association4Tracking}
\end{abstract}

}

\maketitle

\IEEEdisplaynontitleabstractindextext
\IEEEpeerreviewmaketitle

\IEEEraisesectionheading{\section{Introduction}\label{sec:introduction}}


\noindent With the rapid proliferation of consumer cameras, events such as surprise parties, group games and sports events, are increasingly being recorded from multiple views. The challenges in tracking and reconstructing such events include: (a) large scale variation (close-up and distant shots), (b) people going in and out of the fields of view many times, (c) strong viewpoint variation, frequent occlusions and complex actions, (d) clothing with virtually no features or clothing that all look alike (school uniforms or sports gear), and (e) lack of calibration and synchronization between cameras. As a result, tracking methods (both single~\cite{zhang2008global,shitrit2011tracking,choi2015near} and multi-view~\cite{berclaz2011multiple,liem2014joint,rozantsev2017flight}) that rely on motion continuity produces short tracklets. In contrast, tracking-by-association methods relying on pretrained descriptors~\cite{assari2016human,yu2016solution} fail to bridge the domain gap between training data captured in (semi-)controlled environments and event videos captured ``in the wild''.

We present a novel self-supervised person association framework that integrates short-term tracking using motion continuity and long-term tracking using appearance descriptor matching to overcome both their limitations. We show that even a state-of-art pretrained person appearance descriptor is not sufficient to discriminate different people over a long duration and across multiple views. We bridge the domain gap by refining the pretrained descriptor to the event videos of interest with \emph{no} manual interventions (such as manual labeling) and discriminatively learn a robust person association descriptor. Our insight to self-supervision is to exploit three \emph{basic} sources of information in the target domain: (a) short tracklets from tracking-by-continuity methods, (b) multi-view geometry constraints, and (c) mutual exclusion constraints (one person cannot be at two locations at the same time). These constraints allow us to define losses~\cite{chopra2005learning,schroff2015facenet} on triplets of people images -- two of the same person and one of a different one. Even using the most conservative definition of constraint satisfaction (very short tracklets, strict thresholds on the distance to epipolar lines) allows us to \emph{automatically} generate millions of training triplets for domain adaptation. 

While the above domain adaptation stage improves the descriptor discriminability of people with similar appearance, it could also lead to strong semantic bias for people rarely seen in the videos. We address this problem by jointly optimizing the descriptor discrimination on the large labeled corpus of multiple publicly available human re-Identification (ReID) datasets and the unlabeled domain videos using a multitask learning objective. Empirically, the proposed descriptor learning enables easier multi-view association of individual detections or tracklets via clustering. We show that even a simple clustering algorithm such as k-means is sufficient given a known number of people. In practice, since the number of people is unknown, we adopt the continuous clustering framework~\cite{shah2017robust} and enforce soft spatiotemporal constraints from our mined triplets during the construction of the clustering connectivity graph. Since the association is solved globally, there is no tracking drift.

\begin{table*}[]
\centering
\begin{tabular}{|c|c|c|c|}
\hline
Scene         
&\begin{minipage}{.25\textwidth}
\includegraphics[width=\linewidth]{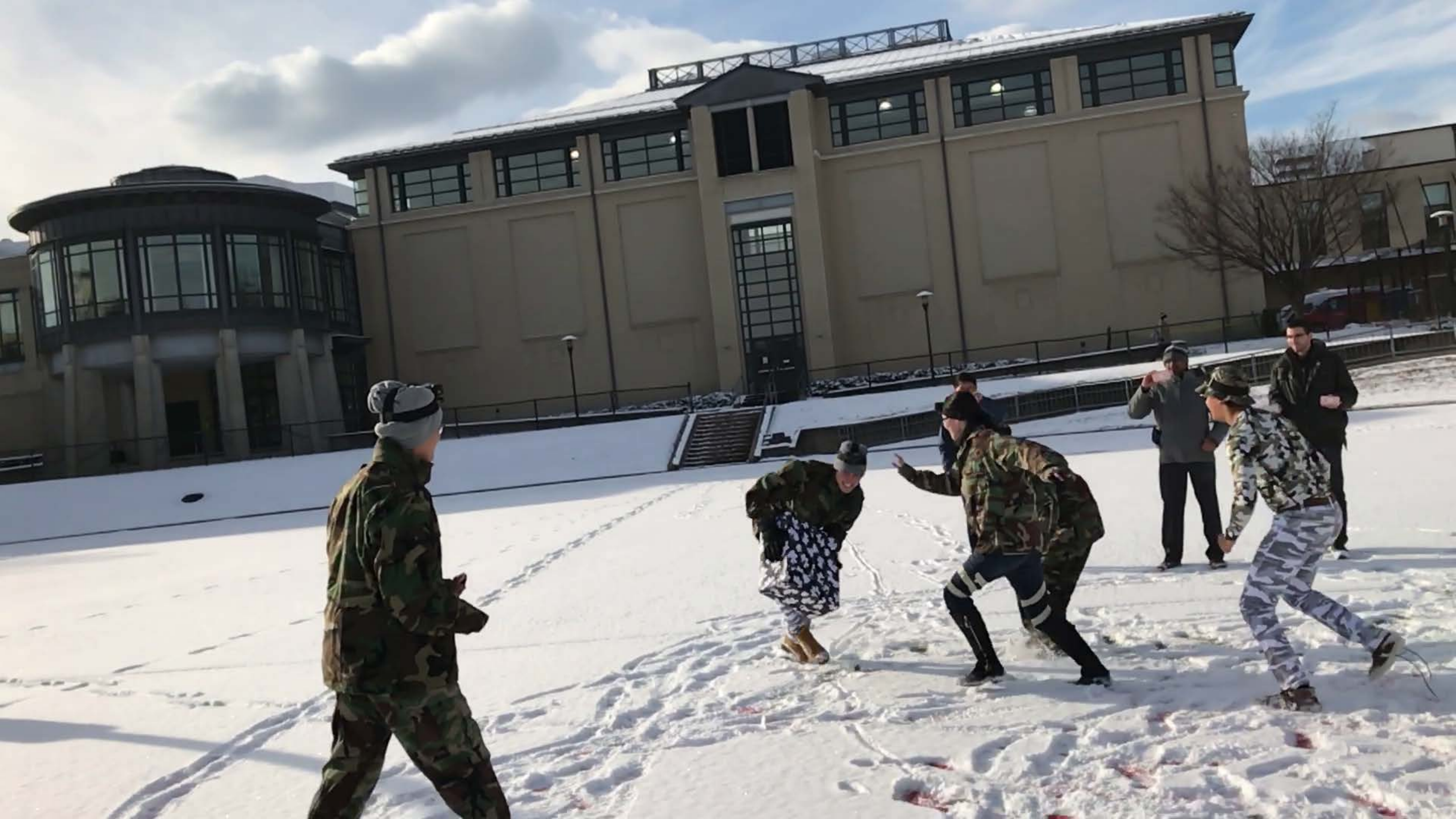}
\end{minipage} 

& \begin{minipage}{.25\textwidth}
\includegraphics[width=\linewidth]{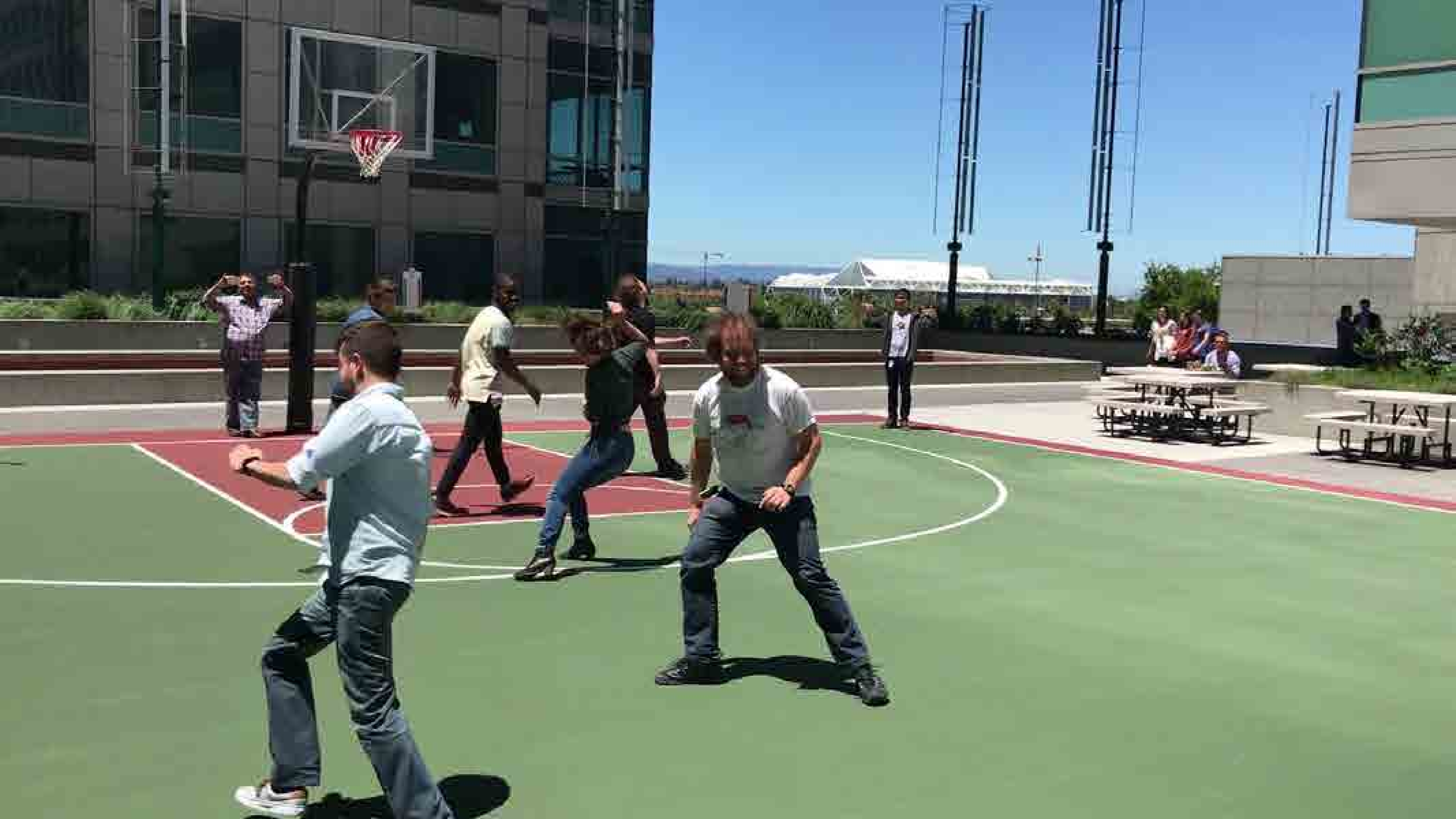}
\end{minipage}

&\begin{minipage}{.25\textwidth}
\includegraphics[width=\linewidth]{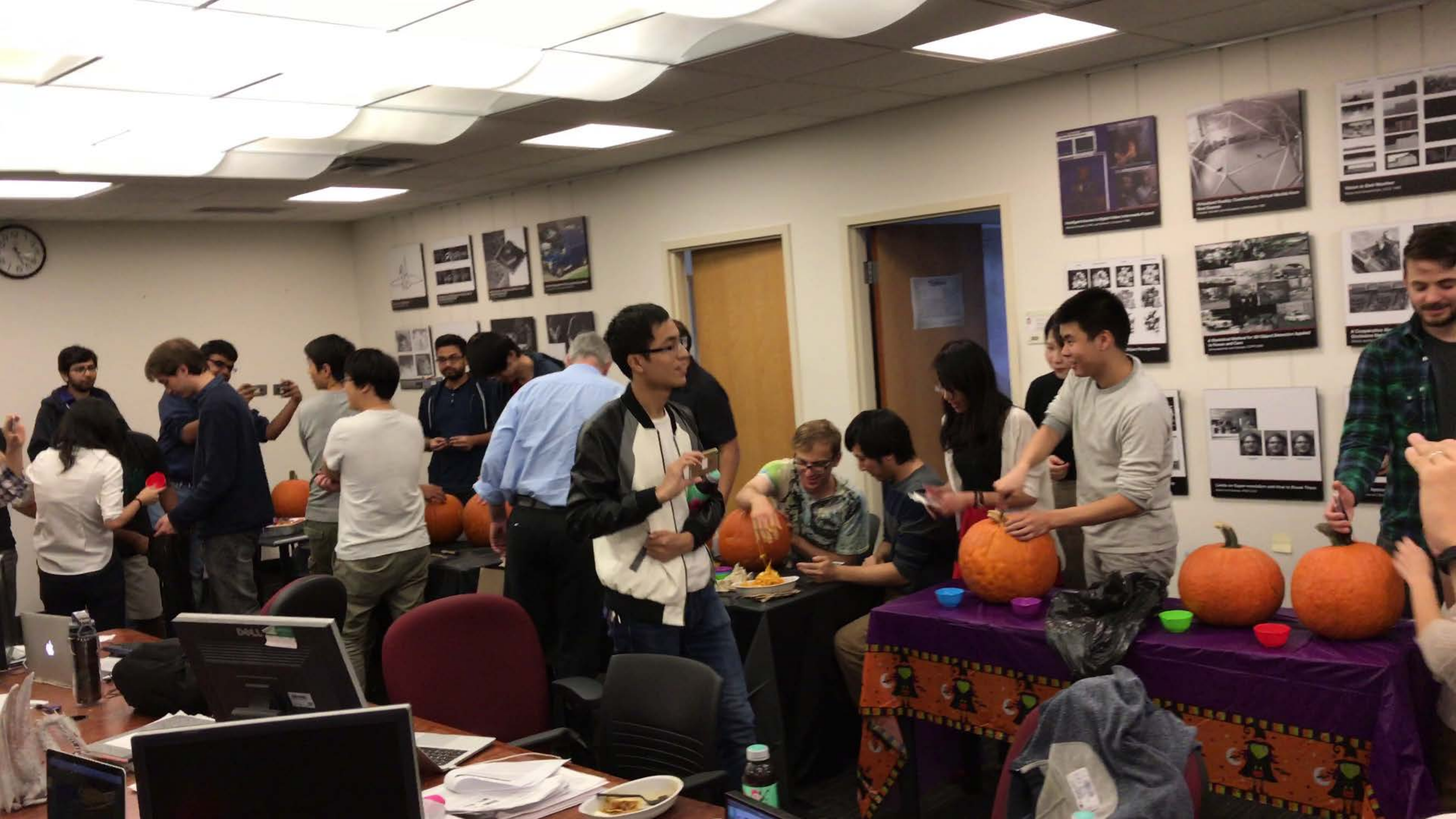}
\end{minipage}      \\ \hline
\# cameras     & 6 head-mounted + 12 hand-held &  16 hand-held &  9 hand-held \\ \hline
Video stats.  &  1920$\times$1080, 60fps, 30s &   1920$\times$1080, 60fps, 60s  &   3840$\times$2160, 30fps, 120s \\ \hline
\# people      &  14                           &  14           &  60           \\ \hline
Tracklet noise &  2\%                         &  11\%         &  3\%          \\ \hline
\end{tabular}	
\caption{Three new group activity tracking datasets scenes: Chasing [C] (left) , Tagging [T] (middle), and Halloween [H] (right).  [C] has 6 people with camouflage and 3 others with dark clothing and cannot be distinguished without attention to detail. Most people in [T] wear feature-less clothing making feature-tracking hard. [H] is from an actual surprise birthday during the Halloween party and suffers from significant motion blur. The scene and camera behavior for these sequences were not staged. Tracklet noise is the percentage of tracklets with at least two people grouped into a single track.}
\label{Tab:SceneStats}
\end{table*}

We validate our framework on the recent WILDTRACK dataset~\cite{ETHZ} as well as three new challenging datasets of complex and highly dynamic group activity: Chasing [C], Tagging [T], and Halloween [H], captured by up to 18 mobile cameras (see Tab.~1). We show significant accuracy improvement in people association over the state-of-art pretrained human ReID model (18\% for [C], 9\% for [T], and 9\% for [H]).


To further demonstrate the impact of the improved descriptor, we use our association to drive a complete pipeline for 3D human tracking to estimate spatially stable and temporally coherent 3D skeleton for each tracked person. Compared to the baseline, our method shows significant improvement (5-10X) in 3D skeleton reconstruction and stability, minimizing tracking noise. We believe that this demonstrates, for the first time, stable and long duration 3D human tracking in actual chaotic live group events. 

We leverage the tracked 3D human skeletons to merge multiple video streams into a multi-angle video by selecting video chunks where the visible person is most frontal to the viewing camera. Such a multi-angle cut is needed because it is unlikely that any person in the scene is clearly visible in a single video stream for a complex group activity event. Our cut algorithm detects inter-human occlusions and determines the appropriate moment to switch the camera while still maintaining the flow of the action well. This system provides an easy visual interface for people tracking from multiple cameras of crowded activities.

\noindent \textbf{Contributions:} (1) We present a simple but powerful self-supervised domain adaptation of person appearance descriptor framework using monocular motion tracking, mutual exclusive constraints, and multi-view geometry \emph{without} manual annotations. (2) We demonstrate that the discriminative appearance descriptor allows a reliable association via simple clustering. This advantage enables a first-of-a-kind accurate and consistent markerless motion tracking of multiple people participating in a complex group activity from mobile cameras \emph{``in the wild"}, with further application to multi-angle video for intuitive tracking visualization. These contributions are \emph{orthogonal} to advances in ReID descriptor, single-view tracking (tracklet), or clustering algorithms---any of these could be starting points for our method. (3) We introduce three challenging datasets with labeled people association for markerless motion capture. 

\section{Related Work}

\noindent Our work is related to the themes of people Re-Identification (ReID) and multi-view motion tracking. People ReID focuses on learning descriptors that match people across views and time. Recent advances can be attributed to large and high-quality datasets~\cite{li2014deepReID,zheng2016mars,ristani2016MTMC}, and strong end-to-end descriptor learning. Common approaches include verification models~\cite{li2014deepReID,ahmed2015improved,cheng2016person}, classification models~\cite{wu2016enhanced,xiao2016learning}, or their combinations~\cite{sun2014deep,McLaughlin2016videoReID}. Some recent works also consider body part information~\cite{li2017learning,zhao2017spindle} for fine-grained descriptor learning. We adopt similar models~\cite{xiao2016learning,sun2014deep} but show how a previous generic person descriptor trained on \emph{labeled} data is insufficient for the reliable human association on the multi-view videos captured in the wild. Our key idea is to automatically exploit basic constraints available in the testing scene itself to adapt the person descriptor with \emph{no} manual annotations. Thus, our model is event (scene) specific rather than being a generic human ReID model.

People tracking approaches formulate person association as a global graph optimization problem by exploiting the continuity of object motion (tracking-by-continuity); examples include~\cite{zhang2008global,milan2014continuous,dehghan2015gmmcp} for single view tracking, and~\cite{berclaz2011multiple,shitrit2014multi,wang2016tracking} for multi-view tracking from surveillance cameras. These approaches use relatively simple appearance cues such as the histogram of color, optical flow, or just the overlapping bounding box area~\cite{shitrit2011tracking,choi2015near,nguyen2004fast,alt2010rapid,collins2012multitarget,butt2013multi} for monocular settings or 3D occupancy map from multi-view systems~\cite{liem2014joint,fleuret2008multicamera,wu2012coupling}. These methods aim to generate reliable short-term tracklets as the targets permanently disappear after a short time. We tackle people tracking in recurrent scenes and our algorithm takes those single-view tracklets as inputs to produce their associations for the entire event. Additionally, whereas existing multi-view tracking algorithms are limited to fixed and synchronized cameras~\cite{liem2014joint,fleuret2008multicamera,wu2012coupling,baque2017deep}, our framework is applicable to uncalibrated moving cameras and can temporally align multiple videos automatically.

Recently,~\cite{dehghan2015target,milan2015joint,yu2016solution,Tang17Tracking} combine global graph optimization and discriminative appearance descriptors and show clear improvements over isolated approaches. The closest to our work is Yu at al.~\cite{yu2016solution}. However, their method assumes \emph{known} number of people captured in controlled settings and solve a challenging $L_0$ optimization using a sophisticated solution path approach. We tackle a similar problem but in unconstrained settings with \emph{unknown} number of people and moving cameras using a much simpler clustering algorithm. This is possible because of our discriminative but automatic scene-aware person descriptor.

We use our person association approach to drive a complete pipeline for 3D human tracking. While markerless motion tracking has been widely demonstrated in laboratory setups~\cite{elhayek2012spatio,liu2013markerless,stoll2011fast,joo2015panoptic} and more recently in general settings~\cite{rhodin2015versatile,moreno20173d,pavlakos2017coarse,elhayek2017marconi}, thanks to advances in CNN-based body pose detectors~\cite{cao2016realtime,newell2016stacked}, these methods showcase the results on activity involving 1 or 2 people staying in a constrained area (never have to re-associate people) with minimal interactions (inter-occlusion is not considered). We note the recent related work of Rhodin et al.~\cite{rhodin2019neural} that learns shape and appearance embedding for people association in an unsupervised fashion and uses them for multi-person 3D pose recovery. However, this work requires static cameras and extending it to dynamic capture is not straightforward. In contrast, we show 3D motion tracking of complex group activities of up to 14 people in unconstrained settings captured by up to 11 cameras where people frequently move in and out of the field of view. 

Our application to multi-angle video cut is most related to Arev at. el~\cite{arev2014automatic}. However, our goal is to obtain the best front-facing camera view to track a selected person whereas they aim to capture the recorders joint attention. Moreover, the constraints in choosing the best view are fundamentally different as we have the underlying 3D skeleton models whereas they rely on co-visibility of the camera frustums.

\section{Person Appearance Descriptor}
\noindent Our goal is to learn a robust appearance descriptor extractor  $u_x=f(x)$ of a person image $x$ that is similar for images of the same person and dissimilar for different people regardless of the viewing direction, body deformation, and other factors (e.g., illumination) for our domain (target) videos. We start with an extractor $f(x)$, initially trained on a large labeled corpus of multiple publicly available people ReID datasets, and finetune it using the Siamese triplet loss on triplets of images \emph{automatically mined} from the domain videos with no human intervention. While this finetuning stage improves the descriptor discriminability of people with similar appearance, it could also lead to strong semantic bias for people rarely seen in the videos. We address this problem using a multitask learning objective and jointly optimize the descriptor discriminability on the labeled human ReID datasets and the unlabeled domain videos. We iteratively mine the triplets and retrain the descriptor for several (triplet mining) iterations.

\subsection{Person Appearance Descriptor}

\begin{figure}[t]
\includegraphics[width=\linewidth]{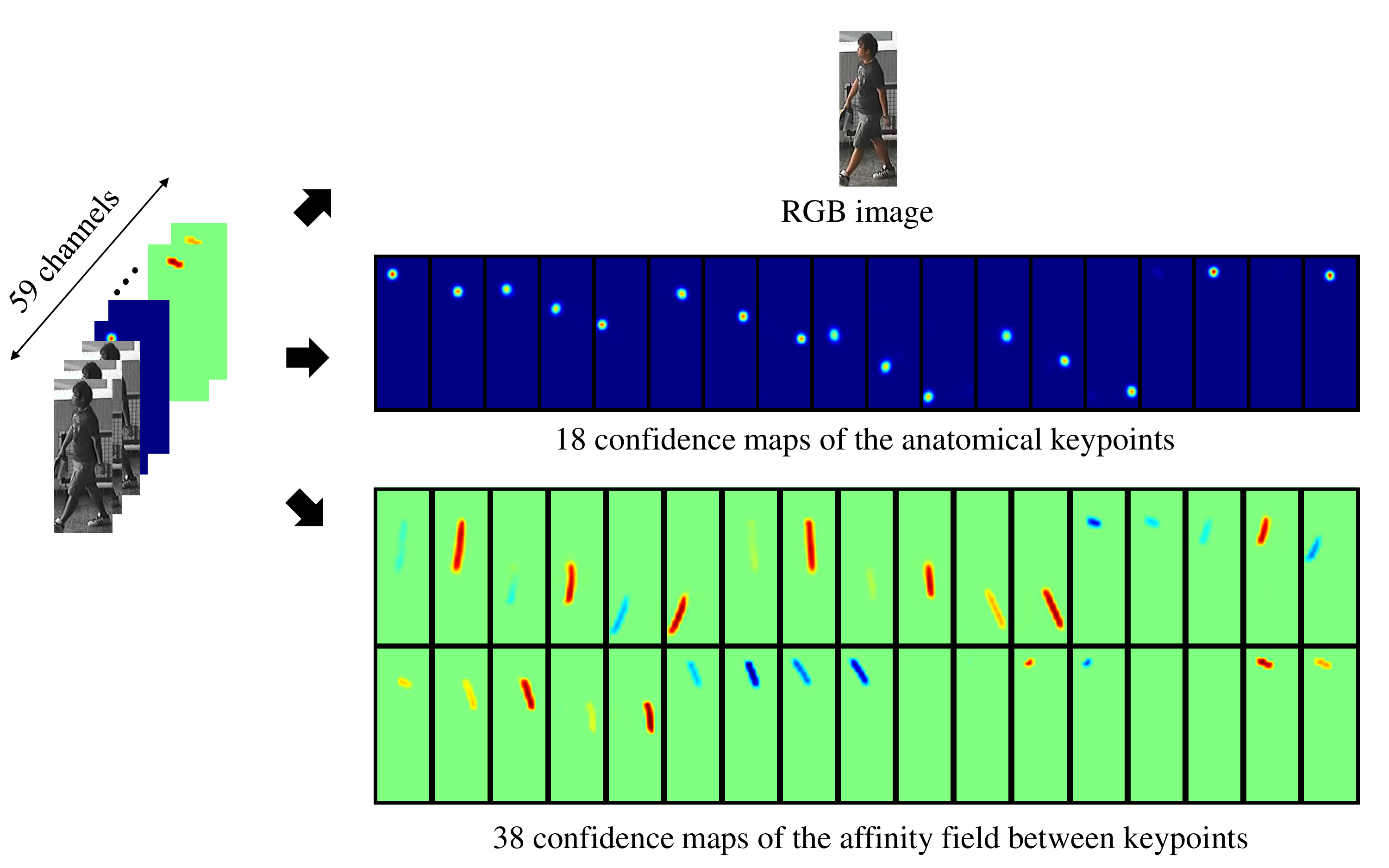}
\caption{The input to our CNN has 59-channels, consisting of the color image, the feature maps of the 18 anatomical keypoints and their affinity fields computed by CPM.}
\label{fig:PoseInvarInput}
\end{figure}

This section describes our pose-insensitive person descriptor extractor $f(x)$. One approach to achieve such invariance is to rectify the input image into a canonical frame~\cite{zhao2017spindle}. However, rectification is problematic due to 2D warping artifacts and wrong pose detection. Instead, we augment the RGB image with the heatmaps of keypoints and their part affinity fields provided by CPM model~\cite{cao2016realtime} (see Figure~\ref{fig:PoseInvarInput}). This representation avoids the viewing direction quantization in rectifying the body parts~\cite{cheng2016person,li2014deepReID} and takes the detection confidence into account to down-weight possible pose detection failures. 

\subsection{Descriptor Adaptation}
\label{STBootstrap}
\noindent Due to possible discrepancies between the appearances of the training sets and our domain application videos, we finetune the $f(x)$ on each of our test video sequences using the contrastive and triplet loss~\cite{chopra2005learning,schroff2015facenet}. The input to our process is triplets of two images of the same person and one image of a different person. We optimize the CNN such that the distance between query and anchor is small and the distance between query and the negative example is large. Our loss function is defined as:

\begin{align} \label{eq:Metric_loss}
L_S(u_i, u_i^{+}, u_i^{-}) &=  \|u_i-u_i^{+}\|_2^2 \nonumber\\&+ \max\big(0,  \|u_i- u_i^{-}\|_2^2- m \big) \nonumber\\&+ \max\big( 0,  \|u_i^{+} -u_i^{-}\|_2^2- m \big), \nonumber \\
L_T(u_i, u_i^{+}, u_i^{-}) &=  \max \big( 0,  \|u_i-u_i^{+}\|_2^2 - \|u_i-u_i^{-}\|_2^2 + m \big), \nonumber \\
L_{ST}(u_i, u_i^{+}, u_i^{-}) &= L_S(u_i, u_i^{+}, u_i^{-})+L_T(u_i, u_i^{+}, u_i^{-}),\nonumber
\end{align}
where, $(u_i, u_i^{+}, u_i^{-})$ is the triplet of two positive and a negative unit norm descriptor, respectively, and $m$ (set to $2$ for all experiments) is the margin parameter between two distances. The total loss function for finetuning is defined as:
\begin{eqnarray}\label{eq:Boostrap_loss}
E_{ST} = \min_{f} \sum_{i=1}^{N_d}  L_{ST}(u_i, u_i^{+}, u_i^{-}),\nonumber
\end{eqnarray}
where, $N_d$ is the number of triplets in the domain videos. We optimize the model using Stochastic Gradient Descent. Empirically, we found hard-negative mining hurts the learning due to possibly erroneous triplets in the first triplet mining stage and thus, only use this trick in later iterations.

\subsubsection{Automatic Triplet Generation}
\label{sec:TMining}
\noindent\textbf{Single-view triplets}: For every video, we first apply CPM to detect all the people and their corresponding anatomical keypoints. Given these detections, we can easily generate negative pairs by exploiting mutual exclusive constraints, i.e. the same person cannot appear twice in the same image. In addition, we can create positive pairs by using short-term motion tracking. We create motion tracklets by combining three trackers: bidirectional Lucas-Kanade tracking of the keypoints, bidirectional Lucas-Kanade tracking of the Difference of Gaussian features found within the detected person bounding box, and person descriptor matching between consecutive frames. The tracklet is split whenever any of the trackers disagree. We also monitor the smoothness of the keypoint 2D trajectories and split the tracklet whenever the instantaneous 2D velocity is $3$ times greater than its current average value. More sophisticated approaches such as~\cite{milan2014continuous,dehghan2015gmmcp} can also be used for better tracklet generation. Images corresponding to the same motion tracklet constitute positive pairs for our finetuning.\\

\noindent\textbf{Multi-view triplets}: We enrich the training triplets with positive pairs across views by using multi-view geometry -- pairs of detections corresponding to a single person in 3D space must satisfy epipolar constraints. Since our videos are captured in the wild, they are unlikely to be synchronized. Thus, we must first estimate the temporal alignment between cameras to use multi-view geometry constraints. Assuming known camera frame rate and start time from the video metadata, which aligns the videos up to a few seconds, we linearly search for the temporal offset with the highest number of inliers satisfying the fundamental matrix. A byproduct of this alignment process is the corresponding tracklets across views, which form our positive training pairs.

More specifically, let $\textbf{k}^n_{i}(t)=\{k_i^{n_{t,1}}, ..., k_i^{n_{t,18}}\}$ be the set of anatomical keypoints of the people detection $n$ at frame $t$ of camera $i$, and $\textbf{T}_i^l = \{n_0, ..,n_F\}$ be a tracklet $l$ containing the images of the same person for $F$ frames. Let $M_c = (\textbf{T}_i^l, \textbf{T}_j^k)$ be the candidate tracklet pair $c$ of the same person, computed by examining the median of the cosine similarity score of between all pairs of descriptors~\footnote[2]{At this stage, the descriptors are extracted using a pretrained CNN.} within the tracklets, for camera pair $(i,j)$ and $\textbf{M}_{i,j}$ be all putative matched tracklets for camera pair $(i,j)$. We set the similarity threshold to $0.5$ and add those candidate matches to the hypothesis pool until their ratio-test threshold drops below $0.7$. We use RANSAC with the point-to-line (epipolar line) distance as the scoring criterion to try all possible time offsets within the window of $[-2W,2W]$ frames to detect the hypothesis with the highest number of geometrically consistent matched tracklets:
\begin{eqnarray}\label{eq:FmatTracklet}
I {\leftarrow} \underset{M_c \in \textbf{M}_{i,j}}{\textbf{RANSAC}}\sum_{w=-W}^W\sum_{t=1}^{F} \sum_{\substack{n=1\\n \in \textbf{T}_i^l(t)\\m = \textbf{T}_j^k(t+w)\\  (\textbf{T}_i^l,\textbf{T}_j^k) \in M_c}}^{N_i(t)} \sum_{p=1}^{18} d(k_i^{n,p},k_j^{m,p},\mathbf{F}_{i,j}(t)),\nonumber
\end{eqnarray}
where, $N_i(t)$ is the number of people detected in camera $i$ at frame $t$, $I$ is the number of inliers, and $d(x_1, x_2, \mathbf{F}_{i,j}(t))$ is the bidirectional point-to-line distance characterized by the fundamental matrix $\mathbf{F}_{i,j}(t)$ between the camera pair. $\mathbf{F}_{i,j}(t)$ can either be estimated by calibrating the cameras with respect to the scene background or explicitly searched for using the body keypoints during the time alignment process. We prune erroneous matches by enforcing cycle-consistency within any triplet of cameras with overlapping field of view. We set $W$ to twice the camera frame rate and use the video start time to limit the search.

\subsection{Multitask Person Descriptor Learning}

\noindent While finetuning the person appearance descriptor exclusively on the test domain could potentially improve discrimination of similar looking people, using it alone may result in semantic drift. The learned descriptor has a strong bias toward frequently observed people. The descriptor of different people who are rarely observed together from a single camera cannot be forced to be different due to the lack of mutual exclusive constraints.

We jointly learn the person descriptor from both the large scale labeled human identity training data and the scene specific (target) videos. Since the model must predict the identity of the person from the labeled datasets, it is expected to output discriminative descriptors for rarely seen people in the domain videos. On the other hand, since we finetune the model on the domain videos, it should also discriminate people in those sequences better than training solely on other datasets. Mathematically, our multitask loss function is defined as:
\begin{eqnarray}\label{eq:Total_loss}
E_D = \min_{f} (1-\alpha) E_{SM}+\alpha E_{ST},\nonumber
\end{eqnarray}
where, $\alpha$ is the scalar balancing the contribution of two learning tasks. $E_{SM}$ is the standard classification loss: 
\begin{equation} 
\label{eq:ReID_softmax}
E_{SM} = \argmin_{f} \sum_i^{N_s} L_{SM} \big(g(f(x_i)),y_i)\big),\nonumber
\end{equation}
where, $N_s$ is the number of training examples in the labeled corpus datasets, $g$ is a linear function mapping the person appearance descriptor, $f(\cdot)$, to a vector of the dimension of the number of people in the training corpus, and $L_{SM}$ is the softmax loss penalizing wrong prediction of the person ID label. We set $\alpha$ equal to 0.5 for all experiments.

\section{Analysis of Human Descriptor Learning}
\begin{table*}[t]
\centering
\begin{tabular}{|c|c|c|c|c|c|c|c|c|}
\hline
\small{name}        & \begin{tabular}[c]{@{}l@{}}\small{patch size/} \\\small stride\end{tabular} & \begin{tabular}[c]{@{}l@{}}\small{output} \\\small side\end{tabular} & \small{\#1$\times$1} & \begin{tabular}[c]{@{}l@{}}\small{\#3X3} \\\small reduce\end{tabular} & \small{\#3x3} & \begin{tabular}[c]{@{}l@{}}\small{double \#3X3} \\\small reduce\end{tabular} & \begin{tabular}[c]{@{}l@{}}\small{double} \\\small \#3X3\end{tabular} & \small{pool+proj} \\ \hline
\small{input}       &                                                              & \small{59x288x112}  &       &              &       &                     &              &                 \\ \hline
\small{conv 0 -- conv 5} & \small{3x3/2}                                                        & \small{32x72x28}    &       &              &       &                     &              &                 \\ \hline
\small{inc 1a} &                                                              & \small{256x72x28}   & \small{64 }   & \small{64}           & \small{64}    & \small{64}                  & \small{64}           & \small{avg+64}          \\ \hline
\small{inc 1b} & \small{stride 2}                                                     & \small{384x36x14}   & \small{64}    & \small{64}           & \small{64}    & \small{64}                  & \small{64}           & \small{max+identity}    \\ \hline
\small{inc 2a} &                                                              & \small{512x36x14}   & \small{128}    & \small{128}           & \small{128}    & \small{128}                  & \small{128}           & \small{avg+128}          \\ \hline
\small{inc 2b} & \small{stride 2}                                                     & \small{768x18x7}    & \small{128}    & \small{128}           & \small{128}    & \small{128}                  & \small{128}           & \small{max+identity}    \\ \hline
\small{inc 3a} &                                                              & \small{1024x18x7}   & \small{256}   & \small{256}          & \small{256}   & \small{256}                 & \small{256}          & \small{avg+256}         \\ \hline
\small{inc 3b} & \small{stride 2}                                                     & \small{1536x9x4}    & \small{256}   & \small{256}          & \small{256}   & \small{256}                 & \small{256}          & \small{max+identity}    \\ \hline
\small{fc7}        &                                                              & \small{256}         &       &              &       &                     &              &                 \\ \hline
\small{fc8}         &                                                              &\small{256}           &       &              &       &                     &              &                 \\ \hline
\small{fc9}         &                                                              &\small{ M}           &       &              &       &                     &              &                 \\ \hline
\end{tabular}
\small{\caption{The structure of our CNN model for person ReID. This model is inspired by the Inception architecture, known for its efficiency and expressiveness. inc and fc stand for Inception and fully connected layers, respectively.}}
\label{tab:cnn_structure}
\end{table*}

\begin{table*}[t]
\small
\begin{center}
\begin{tabular}{lllllll}
\hline\noalign{\smallskip}
\noalign{\smallskip}
Input & CUHK03 &MARS & PRID & iLDS & ViPER & 3dPES\\
\noalign{\smallskip}\hline\noalign{\smallskip}
Baselines & 85.4\cite{geng2016deep} & 77.4~\cite{hermans2017defense}	& 43.6~\cite{paisitkriangkrai2015learning} & 64.6~\cite{xiao2016learning} & \textbf{56.3}~\cite{geng2016deep} & 56.0~\cite{xiao2016learning} \\
Big RGB &91.1 &76.9 &55.0 &84.5 &42.4 &70.0 \\ 
Big RGB+KP &92.8 &\textbf{79.8} &60.0 &84.4 &51.9 &78.0 \\ 
Big RGB+PAF &93.1 &79.4 &59.0 &84.5 &49.7 &79.4 \\
Big RGB+KP+PAF &\textbf{93.7} &\textbf{79.8} &\textbf{62.0} &\textbf{85.2} &\textbf{52.5} &\textbf{78.9} \\ 
\hline\noalign{\smallskip}
\end{tabular}
\end{center}
\caption{Ablative analysis of the pose heatmaps for the top-1 accuracy. Using all the heatmaps generated by CPM yields the best accuracy, albeit modest improvements over the key points (KP) or the part affinity fields (PAF) alone. Except for ViPER, which is a small dataset with strong illumination and viewpoint variations, our method consistently outperforms the baselines by a margin.}
\label{tab:1NN_labeled_Ablative}
\end{table*}

\noindent We validate our method on three challenging new sequences: [C], [T], and [H] (see Tab.~\ref{Tab:SceneStats}). In [T], the camera holders are mostly static and appear in low resolution which does not provide enough appearance variation for strong descriptor learning. It also has many noisy single-view tracklets with different people grouped together due to the lack of texture on the clothing and frequent inter-occlusion. There were no constraints on the camera motion or the scene behavior for any sequences. All the cameras in [C] and [T] are spatially calibrated using the ColMap library~\cite{schonberger2016structure}. Calibration fails for [H] due to human motion which frequently occludes the background and strong motion blur. Note that geometric camera calibration is not required for self-supervised descriptor learning. All we need is the fundamental matrix between views which can be self-calibrated using the detected people keypoints. We manually associate the people in our datasets for quantitative evaluation. 

We first train the generic person descriptor extractor $f(x)$ on a combination of 16 different publicly available ReID datasets: VIPeR~\cite{gray2008viewpoint}, 3DPes~\cite{baltieri20113dpes},  ETH~\cite{schwartz2009learning}, iLIDS~\cite{farenzena2010person}, CAVIARA\cite{cheng2011custom}, PRID\cite{hirzer2011person}, V47\cite{wang2011re}, WARD\cite{martinel2012re}, CUHK02~\cite{li2013locally}, CUHK03~\cite{li2014deepReID}, CUHK04\cite{xiao2017joint}, RAiD\cite{das2014consistent}, Shinpuhkan\cite{kawanishi2014shinpuhkan2014}, MARS\cite{zheng2016mars}, CMU Panoptics studio~\cite{joo2017panoptic}, and SAIVT\cite{bialkowski2012database}. Each dataset was collected with very different locations with different demographics, various camera setups and image resolution. CUHK02 CUHK03, DUKE-MTMC, MARS were captured on campus, where many students wear backpacks. PRID contains pedestrians in street views, where crosswalks appear frequently in the dataset. VIPeR images have significant illumination variation across different camera views. iLIDS was captured at the airport with many people dragging luggage. The CMU dataset, captured in the CMU Panoptics studio, contains strong viewpoint and pose variations. ETH was captured from a single moving camera in street views for multi-target tracking purpose. Notably, CUHK04, contained both images captured around an urban city and movie snapshots, has rich variations of viewpoints, lighting, background conditions. This dataset has the largest number of people (identities) but with very few views for each of them (on average 3 images). For most large-scale datasets (MARS, DUKE\_MTMC), the ground truth person bounding boxes were generated by computer algorithms, which heavily suffers from person misalignment. The resulting labeled training corpus has approximately 200k images of nearly 16k different people.

Tab.~2 shows our detailed CNN model for the extractor $f(x)$. Inspired by the ability to learn multi-scale features and model compactness of the Inception architecture~\cite{szegedy2015going}, our customized CNN model passes an input of size 288x112x59 through six convolution layers, six Inception modules (denoted as inc), and three fully connected layers (denoted as fc). The person descriptor is extracted at $fc8$ layer. We use Batch Norm~\cite{ioffe2015batch} followed by ReLU activation at every layer. We regularize training by randomly switching off $50\%$ of the neurons in the $fc7$ layer during training. We employ the standard softmax loss and train the model from scratch using Stochastic Gradient Descent. We train the model with mini-batch of size 240 for 100 epochs. To train the MTL descriptor, we split samples evenly between unlabeled domain data and the labeled ReID datasets for each training batch and train until 1 epoch of the domain data is reached. 

\subsection{Analysis of Pose-Insensitive Person Descriptor}	

\begin{figure}
\includegraphics[width=\linewidth]{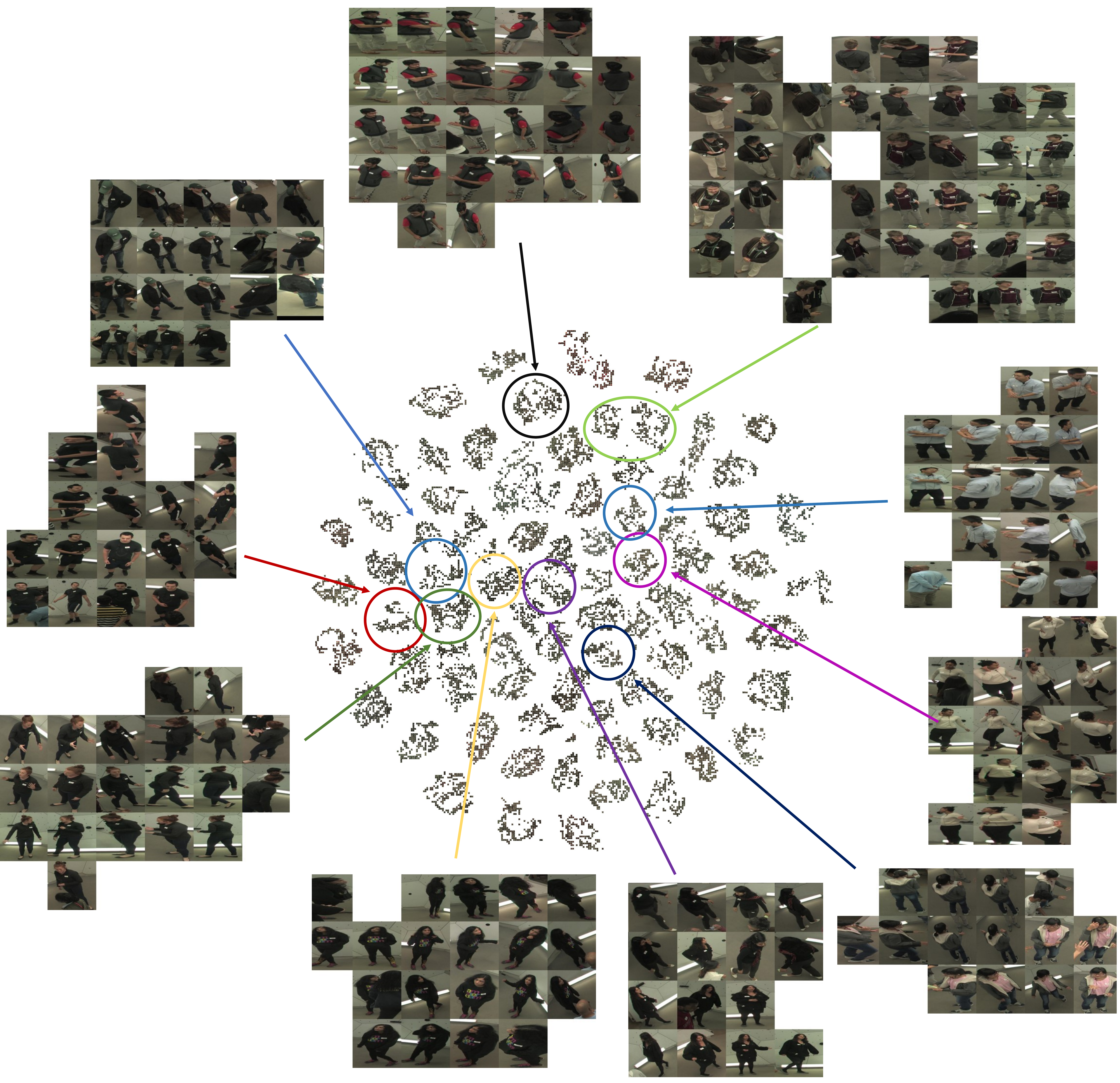}
\caption{The t-SNE visualization of our descriptor for 30k images of 80 people collected by the CMU Panoptic studio. Despite having many people with similar appearances, the images of the same person are clustered together.}
    \label{fig:ViewAndPoseInvariant}
\end{figure}

\begin{figure*}
\centering
\includegraphics[width=1.0\linewidth]{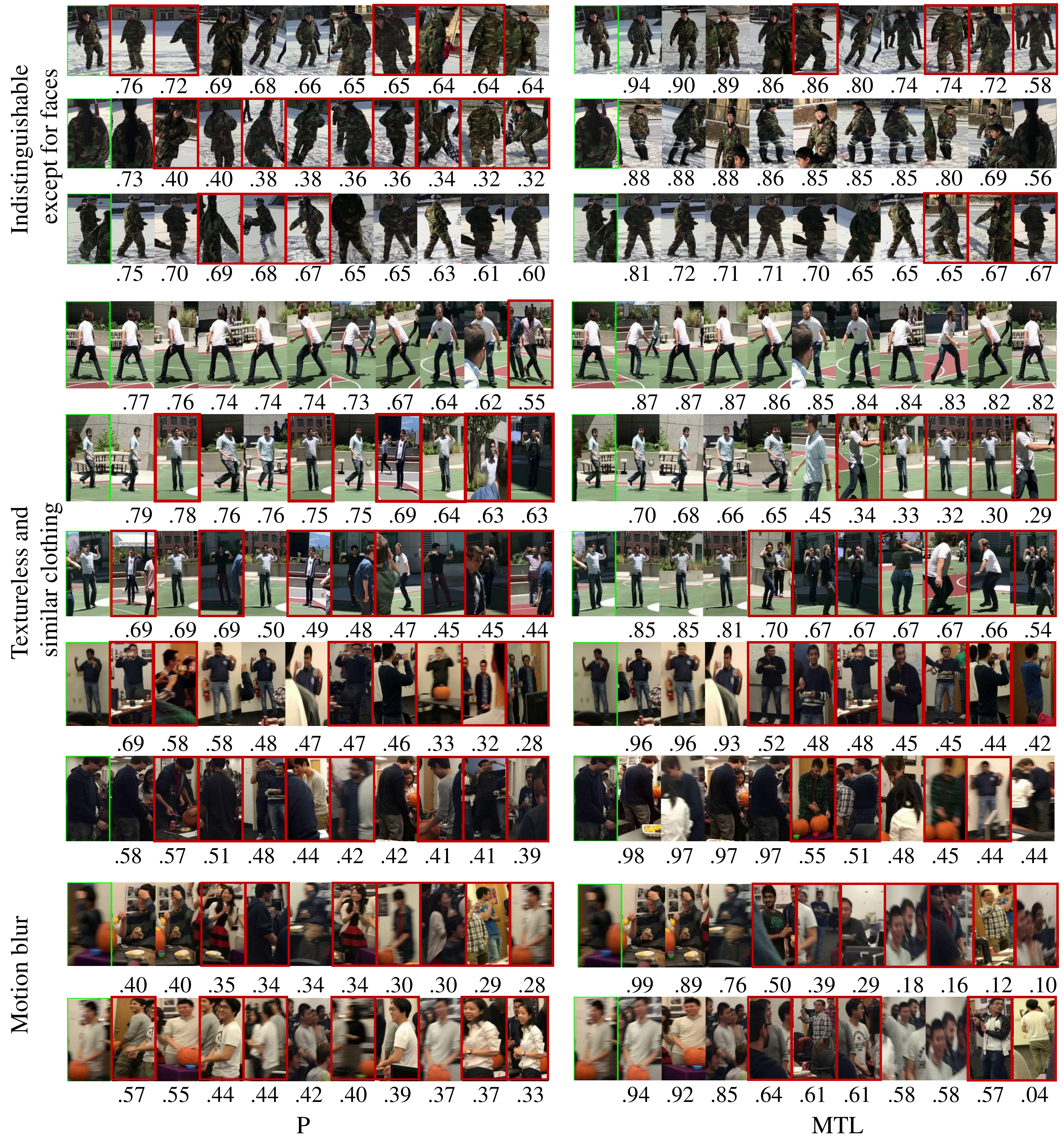}
\caption{10-NN cross-view matching of the several people with confusing appearance and their cosine similarity score using the pretrained model and our multitask descriptor learning (MTL). Green denotes the query and red denotes incorrect matches. We label the query in green and wrong association in red. Our method retrieves more positive matches and provides easy-to-separate similarity score. All top three neighbors are of the same person.}
\label{fig:NN}
\end{figure*}

\begin{figure*}
\centering
\includegraphics[width=1.0\linewidth]{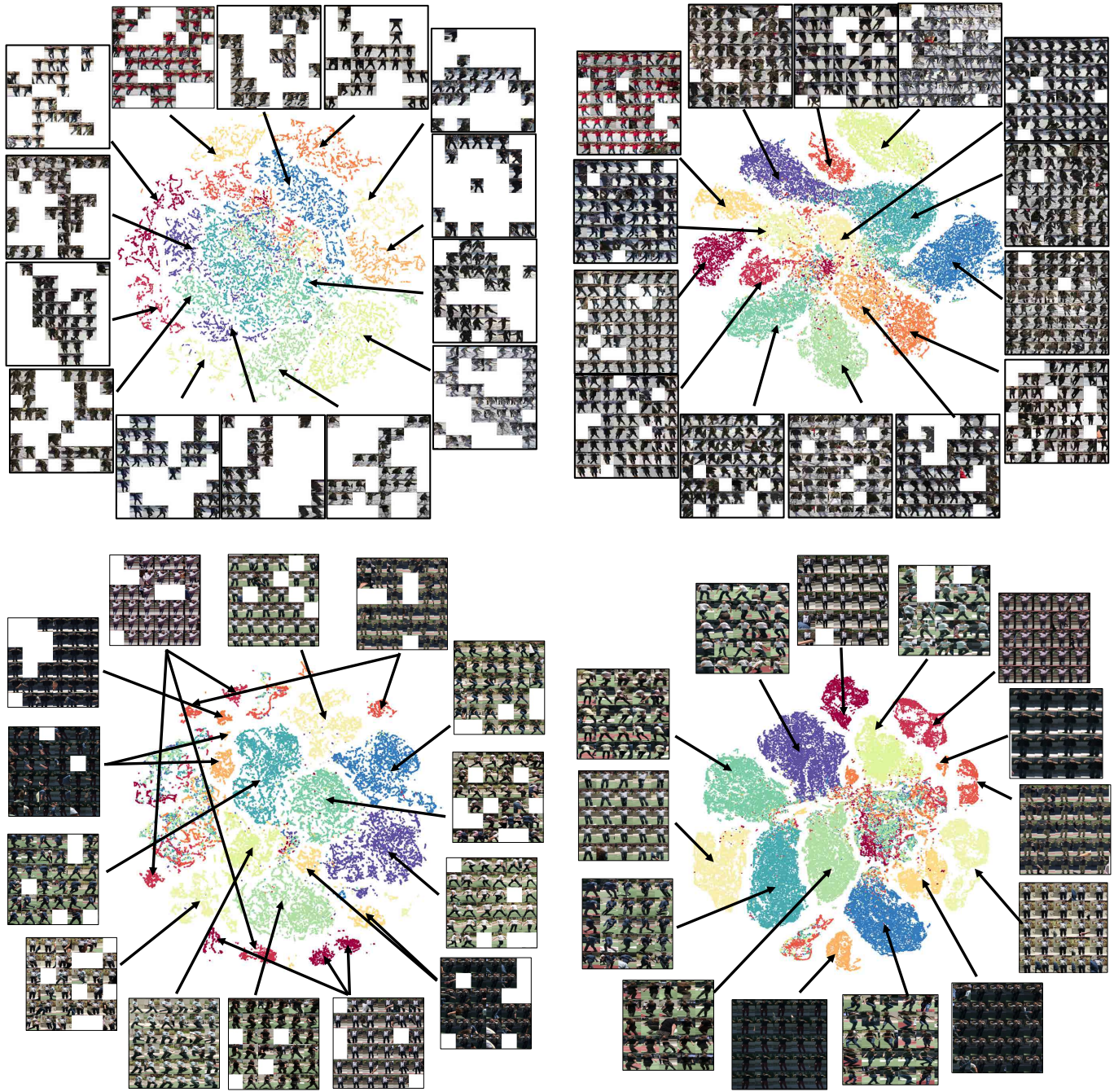}
\caption{t-SNE visualization of the person descriptor extracted using a pretrained model and our  multitask learning (MTL) for sequence [C]. Except for images of the same tracklet within a single view, the pretrained descriptors are scatter. Our descriptor groups images of the same person from all views and time instances into cleanly separated clusters. See Fig.~\ref{fig:Confusion} for extra quantitative evidences.}
\label{fig:tSNE_Tag_Snow}
\end{figure*}

\begin{figure*}
\centering
\includegraphics[width=1.0\linewidth]{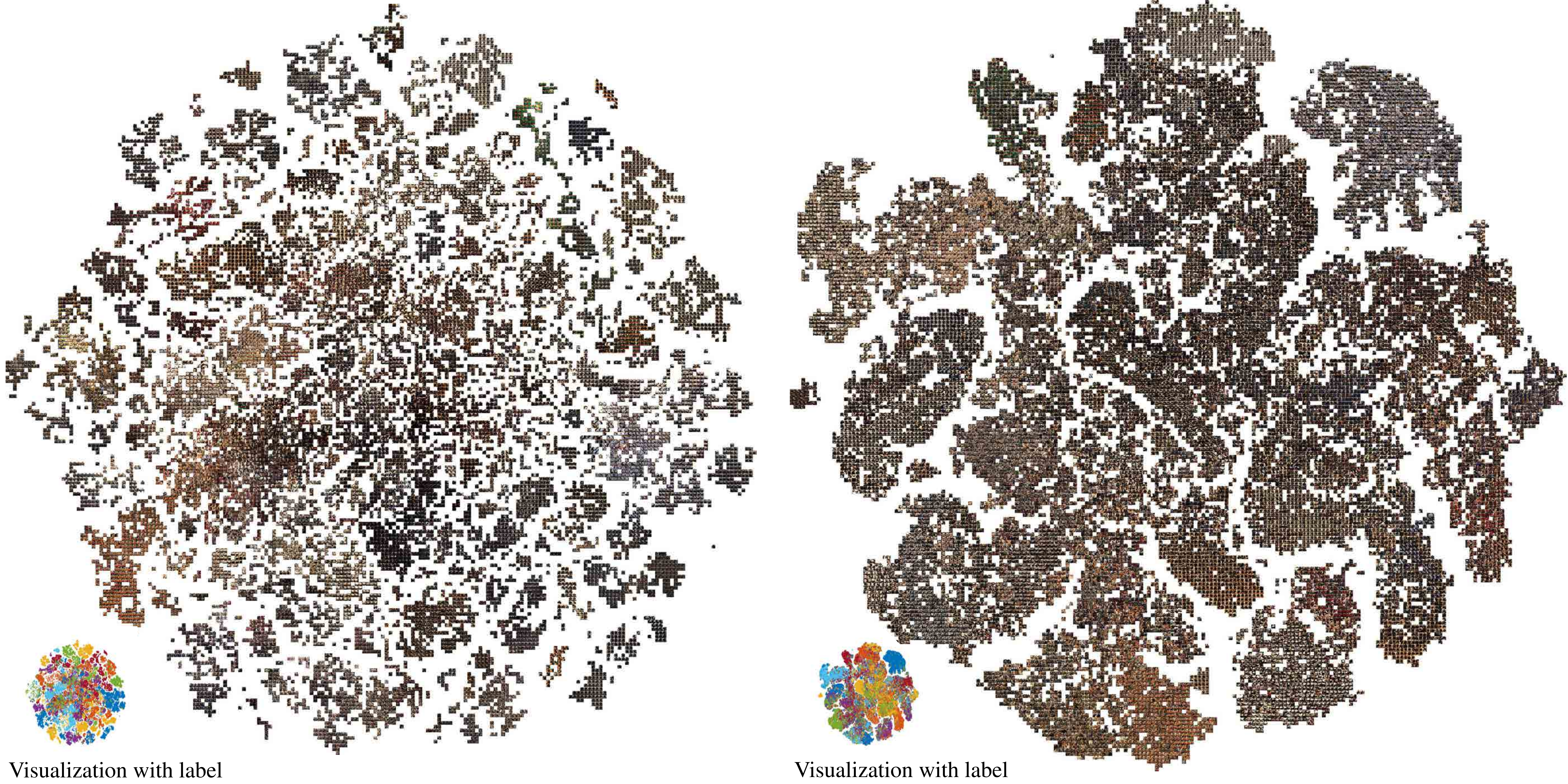}
\caption{t-SNE visualization of the person descriptor extracted using a pretrained model (left) and our  multitask learning model (right) for sequence [H]. Despite being a very complex scene with high number of people, our proposed algorithm shows better discrimination and the same person is better grouped into a single cluster.}
\label{fig:tSNE_Hallo}
\end{figure*}

\boldstart{Qualitative}: We show the pose and viewpoint insensitive properties on the data collected from the CMU Panoptic Studio using t-SNE visualization \cite{maaten2008visualizing} in Fig.~\ref{fig:ViewAndPoseInvariant}. None of the people are the same people as in the CMU dataset used in the training. Despite the similar appearance of multiple people, the images of the same person are clustered together. This shows strong evidence of the pose and viewpoint insensitivity of our descriptor.  

\boldstart{Quantitative}: Tab.~\ref{tab:1NN_labeled_Ablative} shows the comparison between our approach and the recent methods for the top-1 matches on six commonly used datasets and our ablative analysis of how different heatmap categories affects the matching accuracy. For video dataset such as MARS, most methods compute the averaging distance of the learned feature descriptor over all pairs of time instances of the tracklets to match between trajectories. We perform per-frame matching, which is more challenging. Our approach outperforms most other methods by a margin except for ViPER, which is a small dataset with strong variations in viewpoint, image quality, and lighting condition. Since the total number of images from PRID, iLIDS, ViPER, and 3dPES comprises less than $5\%$ of the training images, their appearance statistics is likely to be dominated by larger datasets. This potentially explains for their lower accuracies compared to CUHK03 and MARS. Augmenting the color images with the CPM heatmaps improves the accuracy, among which ViPER is boosted by $10.1\%$. Using both the keypoints and part affinity field heatmaps gives the best accuracy, albeit modest improvement over keypoints or affinity field heatmaps alone.

\subsection{Analysis of Descriptor Adaptation}	

Fig.~\ref{fig:NN} shows 10-NN cross-view matching of images of several people with similar appearance or motion blur for all sequences and their cosine similarity score using the pretrained model (P) and our multitask descriptor learning (MTL). The pretrained model retrieves multiple incorrect matches. Our method is notably more accurate. Our similarity score often has a clear transition between correct and incorrect retrievals. Fig.~\ref{fig:tSNE_Tag_Snow} and Figure~\ref{fig:tSNE_Hallo} show a comparison of the 2D t-SNE embedding~\cite{maaten2008visualizing} between the descriptors using P and our MTL approach. While the descriptors extracted from the pretrained model are scattered, our descriptor groups images of the same person from all views and time instances into cleanly separated clusters. Note that while [H] is a very complex scene with up to 60 people, our discriminative MTL still better clusters images of the same person into a single group.

\begin{figure}
\centering
\includegraphics[width=\linewidth]{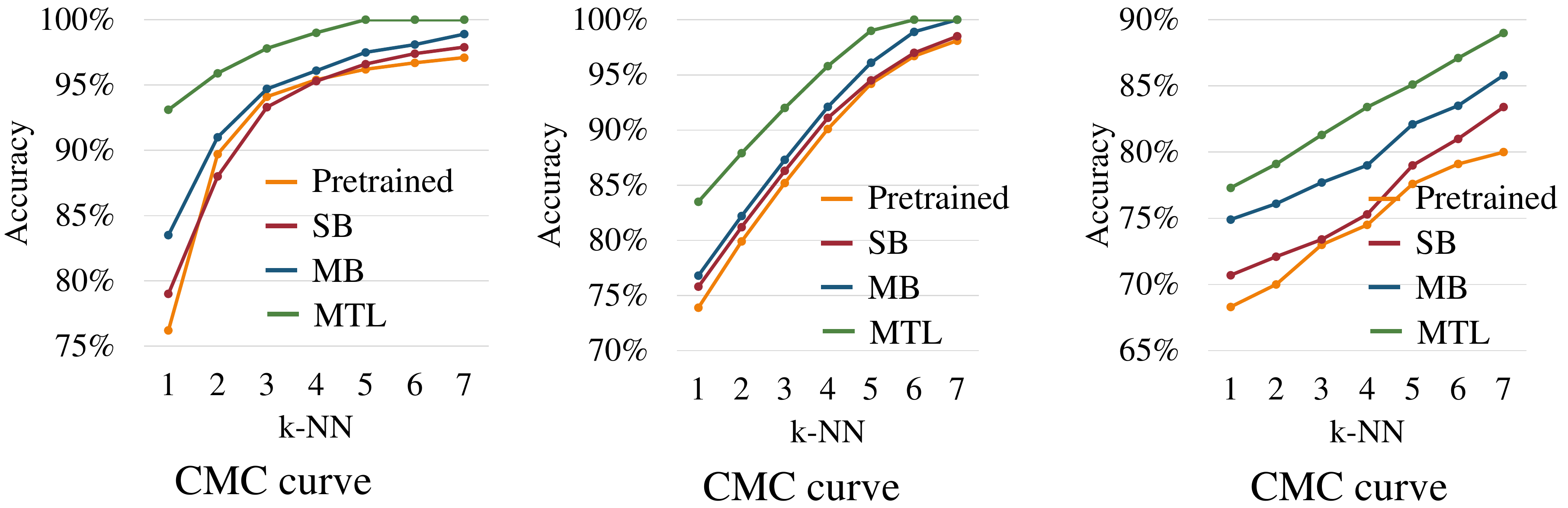}
\caption{The CMC for the Chasing (left), Tagging (middle), and Halloween (right) scene at different stage of our algorithm. Our method outperforms the pretrained model at every stage.}
\label{fig:CMC}
\end{figure}

\begin{figure*}
\centering
\includegraphics[width=.9\linewidth]{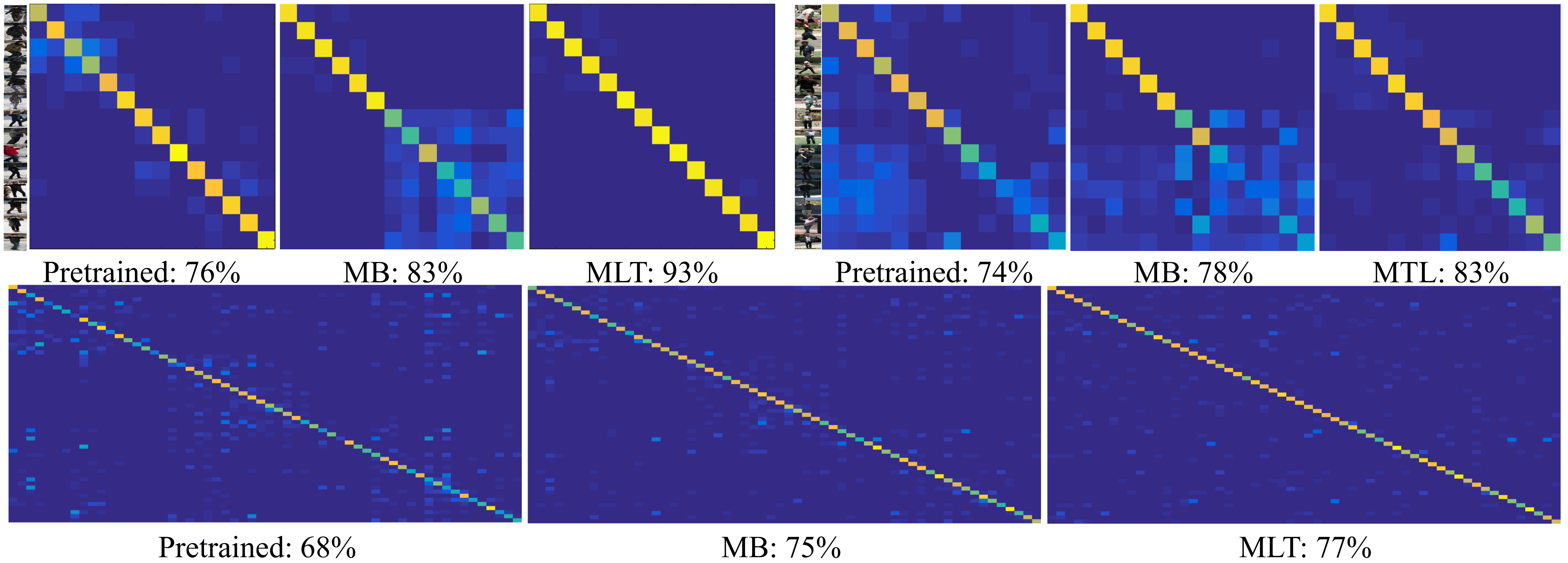}
\caption{The confusion matrix of the top-1 matches for the all sequences ([C] top left, [T] top right, [H] bottom) at different stages: pretrained model, multi-view bootstrapping (MB), and multitask learning (MTL). There are consistent improvements in accuracy as more sophisticated stage is executed.}
\label{fig:Confusion}
\end{figure*}

Fig.~\ref{fig:CMC} shows the Cumulative Matching Characteristic (CMC) for all sequences: Chasing [C], Tagging [T], and Halloween [H]. There are clear improvements over the pretrained model as more sophisticated stages of your algorithm is applied. We further visualize the association accuracy in Fig. \ref{fig:Confusion}. For all sequences, the adapted descriptor improves the discrimination of frequently visible actors: $94\%$ vs. $68\%$ 1-NN classification accuracy for [C] and $90\%$ vs. $75\%$ for [T]. However, the discrimination of descriptor for the camera holders decreases: $56\%$ vs. $85\%$ for [C] and $35\%$ vs. $42\%$ for [T]. Our MTL, combining the strength of the classification and metric learning loss, performs best ($92\%$/$95\%$ for actors/holders on [C] and $89\%$/$61\%$ for [T]) and has an overall baseline improvement of $17\%$ [C], $9\%$ for [T]\footnote[3]{The results for [T] was obtained with cleaned tracklets.}, and $9\%$ for [H]. False matches due to the similar descriptor extracted from the generic CNN model are largely suppressed. 

\begin{figure*}
\centering
\includegraphics[width=.9\linewidth]{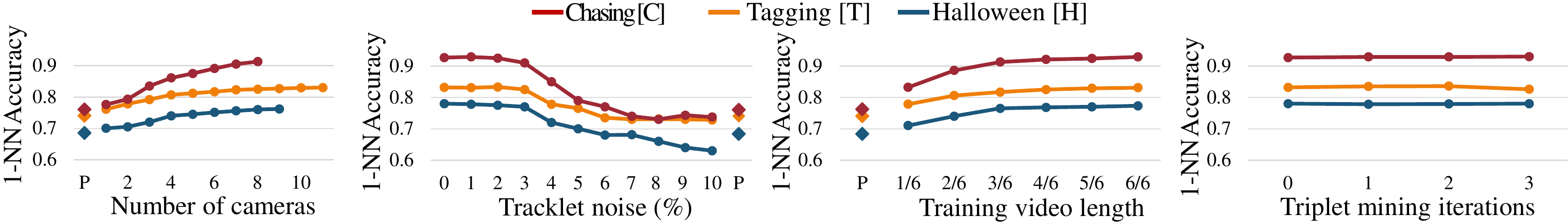}
\caption{1-NN matching accuracy analysis of the proposed method for different number of cameras, percentage of tracklet noise (two or more people grouped in 1 tracklet), fraction of domain data required for generalization, and triplet mining iterations. P denotes the pretrained model. Please refer to the text for the details.}
\label{fig:DescAlb}
\end{figure*}

Fig.~\ref{fig:DescAlb} shows our analysis of the number of cameras, the tracklet noise, the training videos length on 1-NN matching accuracy, and the triplet mining iterations. Multi-view constraints are more helpful than temporal constraints as there are small improvements compared to the pretrained model P when a single camera is used. The algorithm shows noticeable improvement even with as few as 4 cameras.  Yet, for the current scenes, such improvement saturates when more than 6 cameras are used. Regarding tracklet noise, our algorithm can improve the baseline if the noise percentage is less than $4\%$. High noise leads to fewer, and potentially incorrect, multi-view tracklets from pairwise matches and leads to slightly inferior accuracy compared to P. Even finetuning on $1/6th$ of the sequences leads to a notable improvement over P and performance converges after $2/6 th$ of the sequence is used; this indicates that our method could be used on a smaller training set (e.g., first 15 minutes of a game) and applied to the rest. Lastly, we observe marginal accuracy improvement after the 1-st triplet mining iteration. This hints that hard-negative mining is probably not needed and should be avoided at all iterations in our framework.

\section{Applications}

\subsection{Multi-View People Tracking via Clustering}
Using the person descriptor, we cluster detections of the same person across all space-time instances. Since each video contains tens of thousands of detections, jointly clustering all detections for all videos is computationally costly. We adaptively sample the people detector according to their 2D proximity with other detectors and the speed of the detector within each tracklet. All close-by detectors are sampled. Detectors that can be linearly interpolated by others within the same tracklet are ignored. Unreliable detectors with less than 9 keypoints (partially occluded people) detected are also ignored.  

We use the robust continuous clustering framework of Shah and Koltun~\cite{shah2017robust} but explicitly enforce soft constraints from motion tracklets, mutual exclusive constraints, and geometry matching to link detections. Depending on the discrimination of $u$, the correct number of clusters can be automatically determined during the optimization process. This clustering is formulated as the optimization problem:
\begin{eqnarray}\label{eq:Clustering_loss}
C = \min_{\textbf{m}} \sum_{i=1}^N \|u_i-m_i\|_2^2+ \lambda\sum_{(p,q) \in Q}  w_{p,q}\rho(\|m_p - m_q\|_2),\nonumber
\end{eqnarray}
where, $N$ is the number of people detectors, $Q$ is graph connecting data points $u_i$, $\textbf{m} = \{m_1,.., m_N\}$ are the representative of the input descriptors $\textbf{u}$,  $\lambda$ is scalar balancing the maximum curvature between the data and the regularization during the optimization~\cite{shah2017robust}, and $\rho$ is the German-McClure estimator. $w_{p,q} = \frac{\sum_i^N N_i}{N\sqrt{N_pN_q}}$, where $N_i$ is the number of edges connecting $x_i$ in $Q$, balances the strength of the connection $(p,q)$. 

In our settings, the graph $Q$ is mutual $k$-NN graph \cite{brito1997connectivity}. To form $Q$, we first determine the similarity between tracklets by taking the median of the similarity score between all possible person descriptor pairs within the two tracklets. The number of nearest neighbors for each tracklet is chosen such that the distance between different tracklets is 2 times larger than the median of the tracklet self-similarity score. All detectors belonging to the same tracklet are connected with detectors of their $k$ mutually nearest tracklets. We then prune connections that violate the multi-view triplets mined in Section 3.1.1.\\

\begin{table}[]
\centering
\begin{tabular}{c|c|c|c|c|c|c|c|c|c|}
\cline{2-10}
& \multicolumn{3}{c|}{\scriptsize  [C]} & \multicolumn{3}{c|}{\scriptsize [T]} & \multicolumn{3}{c|}{\scriptsize [H]} \\ \cline{2-10} 
& \scriptsize C   &\scriptsize  ARI    & \scriptsize Acc.    & \scriptsize C   & \scriptsize ARI    & \scriptsize Acc.    & \scriptsize C     & \scriptsize ARI     &\scriptsize  Acc.    \\ \hline
\multicolumn{1}{|c|}{\scriptsize ~\cite{shah2017robust}+ P} & \scriptsize 21            & \scriptsize .88    & \scriptsize 90.1      &\scriptsize  66            &\scriptsize  .85    &\scriptsize  86.8      &\scriptsize  86             & \scriptsize .77     &\scriptsize  79.5\     \\ \hline
\multicolumn{1}{|c|}{\scriptsize ~\cite{shah2017robust} + MTL}        & \scriptsize 16            & \scriptsize .97    &\scriptsize  98.3      &\scriptsize  45            &\scriptsize  \textbf{.94}    &\scriptsize  \textbf{95.1}      &\scriptsize  71             & \scriptsize .85     &\scriptsize  88.1      \\ \hline
\multicolumn{1}{|c|}{\scriptsize Kmeans+MTL}        & \scriptsize 16           & \scriptsize \textbf{.98}    &\scriptsize  \textbf{98.7}      &\scriptsize  14            &\scriptsize  .87    &\scriptsize 88.2     &\scriptsize  60             & \scriptsize \textbf{.87}     &\scriptsize  \textbf{88.7}      \\ \hline

\end{tabular}
\caption{Analysis of the clustering algorithms by the number of clusters C, ARI measure and clustering accuracy. Although all methods detected clusters than needed, they are small clusters belonging to the pedestrians who do not participate in the activity (often seen in [T]) or not fully visible bodies due to occlusion. Using ~\cite{shah2017robust} on our (MTL) descriptors performs best, achieving the clustering accuracy of ($98.3$\% for [C] and $95.6$\% for [T]))).}
\label{tab:ClusterConsistency}
\end{table}

\boldstart{Analysis of the Descriptor Benefits}: Tab.~\ref{tab:ClusterConsistency} quantifies the performance of different descriptor learning algorithms by the number of clusters automatically determined by the algorithm, the Adjusted Rand Index (ARI)\footnote[4]{The ARI is a measure of the similarity between two clusters with different labeling systems and is widely used in statistics~\cite{hubert1985comparing}.}, and cluster accuracy for all detected people in both sequences. Using~\cite{shah2017robust} on MTL descriptor performs best. However, for a known number of people, performing the classical K-means clustering on the MTL descriptor also yields comparable accuracy (the precise number of people is only available in [C]). This confirms our descriptor learning as the main driving factor, not the clustering algorithm\footnote[5]{We obtain similar conclusion using the recent ReID descriptor~\cite{luo2019bag}. The quantitative results under the same format as in Tab.~\ref{tab:ClusterConsistency} are as follow: $\{20, 0.90, 90.6\}$, $\{65, .87, 88.1\}$, $\{83, .80, 82.4\}$ for~\cite{shah2017robust}+\cite{luo2019bag} and $\{16, .97, 98.5\}$, $\{44, .95, 95.5\}$, $\{71, .86 88.3\}$ for~\cite{shah2017robust}+\cite{luo2019bag}+MLT.}.
   
\subsection{Markerless Human Motion Capture}

\begin{table*}
\centering	
\begin{tabular}{cc}
\begin{minipage}{0.5\textwidth}
\begin{flushleft}
\begin{tabular}{|l|l|}
\hline
{$E_I(\textbf{K})$}        & {$\sum_{c=1}^C \sum_{t=1}^F \sum_{n=1}^N \sum_{p=1}^{18} \rho\bigg(V_c^{np}(t)\frac{\pi_c(K^{np},t)-k^{np}_c(t)}{\sigma_{I}} \bigg)$}  \\ \hline

{$E_L(\textbf{K},\overline{\textbf{L}})$} & $\sum_{t=1}^F \sum_{n=1}^N\sum_{q\in Q} \bigg(\frac{\overline{L}^{nq} - L_c^{nq}(t)}{\sigma_L}\bigg)^2$ \\ \hline

{$E_S(\textbf{K})$} & {$\sum_{t=1}^F \sum_{n=1}^N\sum_{(l,r)\in S} \bigg(\frac{L_c^{nl}(t) - L_c^{nr}(t)}{\sigma_S}\bigg)^2$} \\ \hline

{$E_M(\textbf{K})$}              & {$\sum_{n=1}^N \sum_{p=1}^{18} \sum_{i=1}^{F-1} \bigg(\frac{K^{np}(i+1)-K^{np}(i)}{\sigma^p_M\Delta(i+1,i)}\bigg)^2$} \\ \hline
\end{tabular}\\	
\end{flushleft}
\end{minipage}&

\begin{minipage}{0.5\textwidth}{
$C$: number of cameras\\
$F$: number of frames\\ 
$N$: number of tracked people\\
$\pi_c(K^p,t)$: projection matrix\\
$V_c^{np}(t)$: visibility indicator\\ 
$L^{nq}(t)$: 3D distance between two points\\
$Q$: keypoint connectivity set\\
$S$: corresponding left and right limb set\\
$\Delta(.,.)$: absolute time differences\\
$\sigma_I$: variation in 2D detection\\
$\sigma_L$: variation in bone length\\ 
$\sigma^p_M$: variation in 3D speed\\
}
\end{minipage}
\end{tabular}
\caption{3D human-aware tracking cost functions.}
\label{tab:3DTracking_cost}
\end{table*}

\begin{figure*}[t]
\centering
\includegraphics[width=1.0\linewidth]{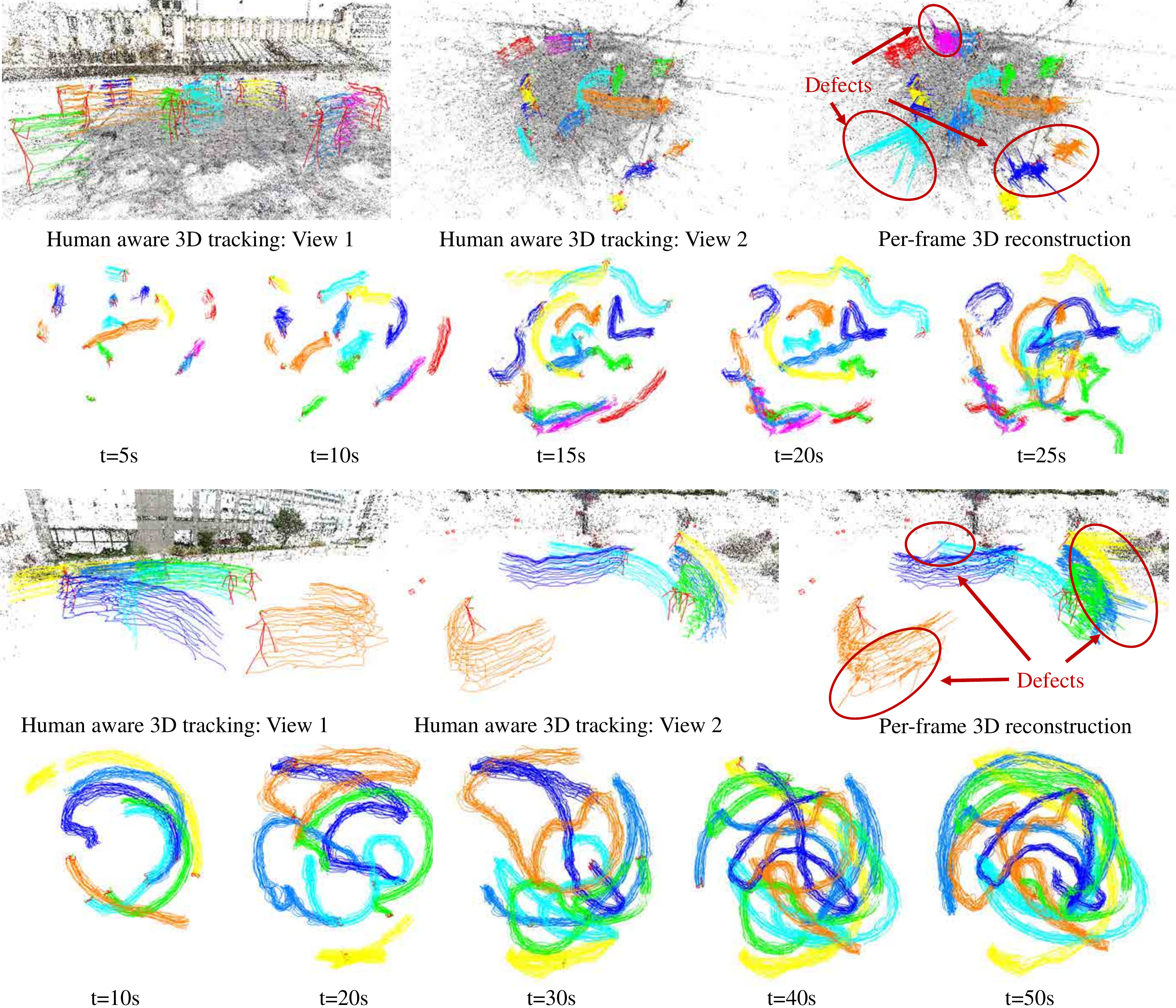}
\caption{3D tracking for [C] (top) and [T] (bottom) for the entire event. Owing to an accurate association, our method gives smooth and clean trajectories despite strong occlusion, similar people appearance, and complex motion pattern. Please refer to the supplementary material for visualization of the comparison with the baseline where obvious tracking artifacts occur.}
\label{fig:SA13D}
\end{figure*}

\label{sec:3DPeoplTracking}
\begin{figure*}[t]
\centering
\includegraphics[width=\linewidth]{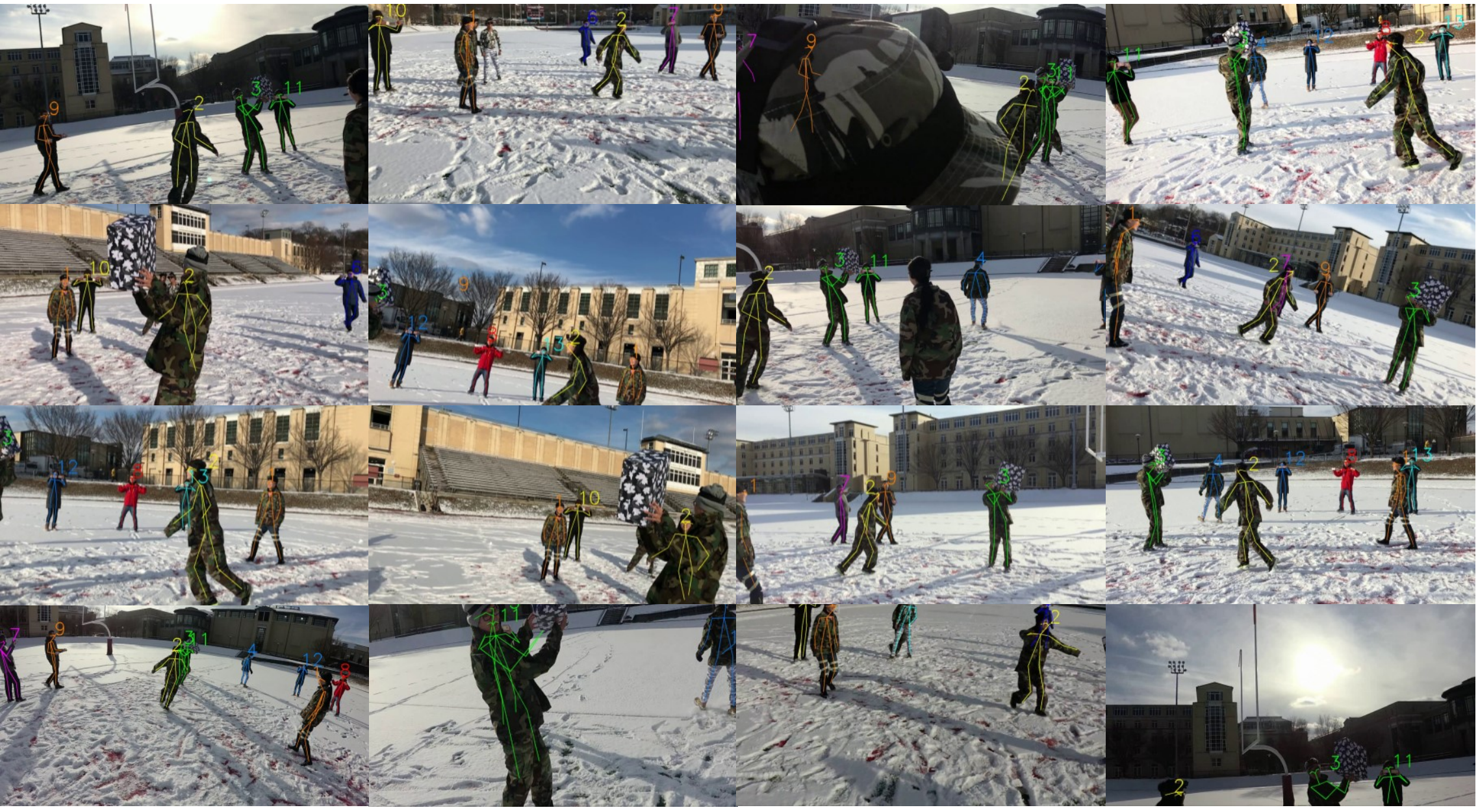}
\caption{The 2D projection of the keypoints to all views corresponds well to the expected person anatomical keypoints and tracks people even through occlusions.}
\label{fig:Snow2D}
\end{figure*}

\begin{table}
\centering
\begin{tabular}{c|c|c|c|c|}
\cline{2-5}
& \multicolumn{2}{c|}{{[C]]}} & \multicolumn{2}{c|}{{[T]}}     \\ \cline{2-5} 
& {Baseline}    & {Ours}  &{Baseline}    & {Ours} \\ \hline	
\multicolumn{1}{|l|}{{Length Dev. (cm)}}      &{8.0}              &{1.5}               &{13.6}              &{1.5}                 \\ \hline
\multicolumn{1}{|l|}{{Symmetry Dev. (cm)}} &{9.1}            &{1.2}               &{10.2}              &{1.4}                  \\ \hline	
\noalign{\smallskip}
\end{tabular}
\caption{Comparison between per-frame 3D skeleton reconstruction using ground truth association and human aware tracking. Temporal integration and the physical body constraints improve the 3D skeleton stability by 5-10X.}
\label{tab:3DTracking}
\end{table}

\begin{figure*}[t]
\centering
\includegraphics[width=\linewidth]{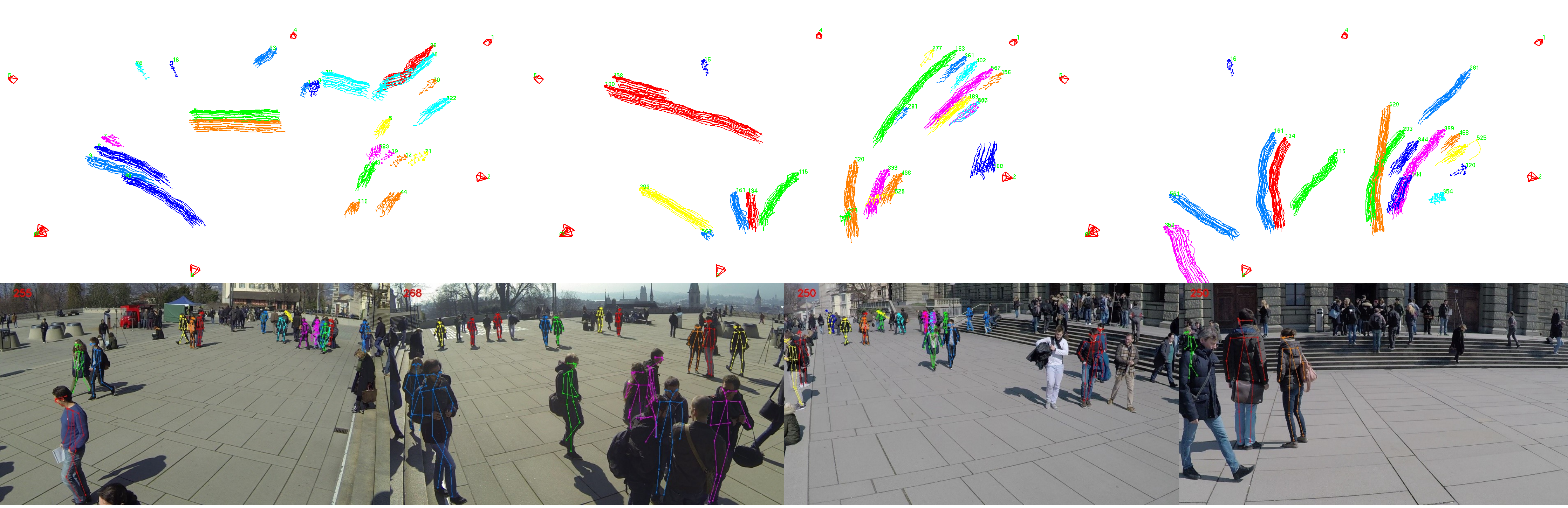}
\caption{3D tracking from 7 cameras of the WildTrack dataset. Owing to an accurate association, our method gives smooth and clean trajectories despite strong occlusion, similar people appearance, and complex motion pattern. Please refer to the supplementary material for visualization of the 3D tracked human. The 2D projection of the keypoints to all views corresponds well to the expected person anatomical keypoints and tracks people even through occlusions.}
\label{fig:WildTrack_3D}
\end{figure*}


We build a pipeline for markerless motion tracking of complex group activity from handheld cameras. We first cluster the descriptors from all camera to obtain person tracking information. For each person (cluster), we wish to estimate a temporally and physically consistent human skeleton model for the entire sequence. This is achieved by minimizing an energy function that combines an image observation cost, motion coherence, and a prior on human shape:
\begin{eqnarray}\label{eq:3D_Tracking}
E(\textbf{K},\overline{\textbf{L}}) =  E_I(\textbf{K}) + E_L(\textbf{K},\overline{\textbf{L}}) +  E_S(\textbf{K}) +  E_M(\textbf{K}),\nonumber
\end{eqnarray}  
where, $\textbf{K}$ is the 3D location of the anatomical keypoints over the entire sequence, $\overline{\textbf{L}}$ is the set of mean limb length for each person. The image evidence cost $E_I$ encourages the image reprojection of the set of keypoints 3D position to be close to the detected 2D keypoints. The human constant limb length cost $E_L$ minimizes the variations of the human limb length over the entire sequence. The left-right symmetric cost $E_S$ penalizes large bone length differences between the left and right side of the person. The motion coherency cost $E_M$ prefers trajectory of constant velocity~\cite{vo2016spatiotemporal}. The formulation for each of these terms are given in Tab.~\ref{tab:3DTracking_cost}. We weight these costs equally.

We initialize $\textbf{K},\overline{\textbf{L}}$ by per-frame RANSAC triangulation of the corresponding person obtained from the person from descriptor clustering and minimize $E(\textbf{K},\overline{\textbf{L}})$ using Levenberg-Marquardt optimizer~\cite{ceres-solver}. Lastly, we fit the SMPL mesh model~\cite{loper2015smpl} to the skeleton to improve the visualization quality.\\

\boldstart{Analysis of the Descriptor Benefits}: As a baseline, we use the ground truth people association to perform a per-frame multi-view triangulation along with limb length symmetry constraints link this reconstruction temporally using ground truth person tracking for visualization. As shown in Fig.~\ref{fig:SA13D}, our method succeeds despite the strong occlusion and complex motion pattern. Please refer to the supplementary material for visualization of the comparison with the baseline where obvious tracking artifacts occur. Quantitatively, we show $5$ to $10$X improvement over the baseline (see Tab.~\ref{tab:3DTracking}). We visualize the reprojection of 3D keypoints to all views for [C] in Fig.~\ref{fig:Snow2D}. The reprojected points are close to the anatomical keypoints. These results validate our algorithm ability to perform accurate markerless motion capture completely in the wild.

\begin{table}[]
\centering
\begin{tabular}{l|c|c|c|c|}
\cline{2-5}
\multicolumn{1}{c|}{}                                & \multicolumn{2}{c|}{\scriptsize{[}C{]}} & \multicolumn{2}{c|}{\scriptsize{[}T{]}} \\ \cline{2-5} 
\multicolumn{1}{c|}{}                                & \scriptsize ARI        & \scriptsize Acc.        & \scriptsize ARI        & \scriptsize Acc.        \\ \hline
\multicolumn{1}{|l|}{\scriptsize \cite{yu2016solution}+P+known \#cluster}           & \scriptsize .82        & \scriptsize 85.3\%          & \scriptsize .76        & \scriptsize 78.5\%          \\ \hline
\multicolumn{1}{|l|}{\scriptsize \cite{yu2016solution}+MTL+known \#cluster}                   &\scriptsize .89        &\scriptsize 91.7\%          &\scriptsize .81        &\scriptsize 82.4\%          \\ \hline
\multicolumn{1}{|l|}{\scriptsize \cite{yu2016solution}+P+GT 3D location+known \#cluster} & \scriptsize .88        & \scriptsize 90.3\%          & \scriptsize .84        & \scriptsize 84.1\%          \\ \hline
\multicolumn{1}{|l|}{\scriptsize \cite{yu2016solution}+MTL+GT 3D location+known \#cluster}        &\scriptsize .96        &\scriptsize 98.2\%          &\scriptsize .87        &\scriptsize 89.8\%          \\ \hline
\multicolumn{1}{|l|}{\scriptsize Ours:MTL+unknown \# clusters}                            &\scriptsize .97        &\scriptsize 98.3\%          &\scriptsize .94        &\scriptsize 95.6\%          \\ \hline
\end{tabular}
\caption{Analysis of multi-view 3D tracking. Due to the inaccurate sparse association between detections (tracklet noise in [T]), wrong number of people (in [T]), and early commitment to perframe 3D human position estimation (sensitive to errors due to wrong/missing inliers in RANSAC),~\cite{yu2016solution} is not as competitive as ours. Discriminative descriptor learned by MTL outperforms pretrained P regardless of the association algorithms. GT 3D is the per-frame estimated location with ground truth association.} 
\label{Tab:SolutionPath}
\end{table}

\begin{table}[t]
\centering	
\label{Tab:CompOthers}
\begin{tabular}{|c|c|c|c|}
\hline
$r$ &  \cite{baque2017deep} &  Ours: No tracklets &  Ours: Full \\ \hline
0.3   & 67.0\%  & 71.6\% &  71.9\%    \\ \hline
0.5   & 74.1\%   & 75.8\% &76.2\%    \\ \hline
\noalign{\smallskip}
\end{tabular}
\caption{MOT-Accuracy comparison for different threshold radius $r$ on WILDTRACK~\cite{ETHZ}. Due to large number of negative samples, our method outperforms~\cite{baque2017deep} even without using single-view tracklet for triplet generation. We observe modest gain in our full method because frequent occlusions and frame sub-sampling prohibit long single-view tracklets.}
\label{Tab:Widltrack}
\end{table}

We also compare our tracking via clustering approach to current arts in multi-view people tracking of~\cite{yu2016solution} on [C] and [T] (results on [H] is not possible due to calibration failure) in Tab.~\ref{Tab:SolutionPath} and with~\cite{baque2017deep}, the current best published 3D tracker on the WILDTRACK\cite{ETHZ} dataset, the current most challenging multi-view tracking dataset in Tab.~\ref{Tab:Widltrack}\footnote{Due to strong inter-person occlusion, many correctly associated detection are seen by less than 3 cameras, which are discarded by our 3D reconstruction algorithm. Thus, we only show the MOT-Accuracy for reconstructed 3D skeletons.}. In both cases, we observe clear improvements using our MTL descriptors. Note that while previous methods require \emph{fixed} cameras (in ~\cite{baque2017deep}) and \emph{known} number of people (in ~\cite{yu2016solution}), 
our algorithm can perform long term tracking of group activities in the wild without such requirements. We show our 3D tracking and the projected skeleton to observed images for the WILDTRACK dataset in Fig.~\ref{fig:WildTrack_3D}. We observe that the 2D projection of the keypoints to all views corresponds well to the expected person anatomical keypoints and tracks people even through occlusions.

\subsection{Semantic Cut for Multi-angle Video}

\begin{figure*}[t]
\centering
\includegraphics[width=\linewidth]{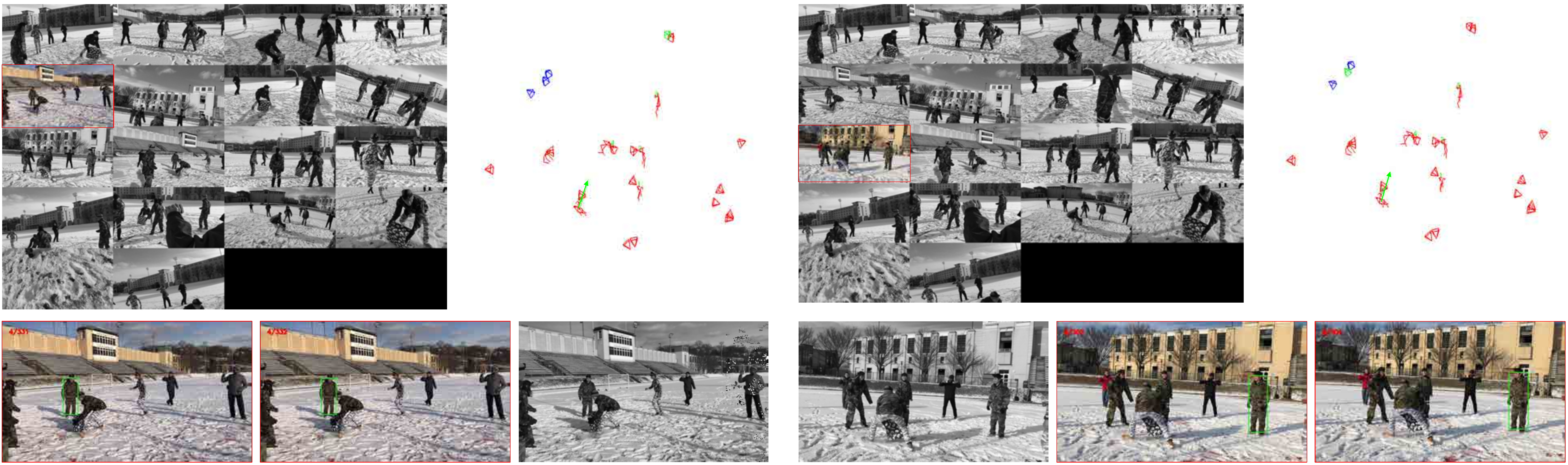}
\caption{A visualization of two shots cut created by our algorithm for the Chase sequence. The red square indicates the selected camera (frame). The top row shows the input images from all camera at a particular time instance and the corresponding 3D view. The tracked person is highlighted in the green bounding box. The green arrow shows his front-facing direction. All visible cameras are shown in blue and the selected camera is shown in green. The bottom row shows consecutive frames of the final video. The non-selected frames are shown in grayscale. Our algorithm detects and switches the camera upon inter-human occlusion.}
\label{fig:DirectCut_Snow}
\end{figure*}

For any complex group activity events, it is unlikely that any person in the scene is well recorded by one video stream. We leverage track3d 3D human skeleton and merge multiple video streams into a multi-angle video by selecting video chunks where the selected person is visible and most frontal to the camera. Similar to Arve et al.~\cite{arev2014automatic}, we model the selection of the camera on a trellis graph and seek for the smallest cost path traversing the graph. The nodes in this graph are frames of camera $c$ and edges are the connection of all consecutive frames from all cameras. \\

\boldstart{Node cost:} This cost is determined by the normal vector of the person torso $n(K)$ and the camera $c$ viewing direction $d_c$ and the distance of the projection of the skeleton $K$ to image center and is written as:
\begin{eqnarray}\label{eq:Cut_Node}
E_n(c,f) = V_c(f) \bigg( \lambda_1 n(K).d_c + \lambda_2 g(|\pi_c(K,f) - c_c|^2, \tau)\bigg) ,\nonumber
\end{eqnarray} 
where $V_c(f)$ is a binary visibility indicator, $\pi_c(f)$ is the camera projection matrix, $c_c$ is the 2D image center location, $g(.,\tau)$ is a thresholding function, only penalizes the cost if it exceeds $\tau$, and $(\lambda_1, \lambda_2)$ are the weights between two error terms.\\

\boldstart{Edge cost:} This cost function is a weighted combination of constant cost $\gamma$ penalizing rapid camera switching and cut-on-action cost penalizing camera switching during action. We determine this cost by the instantaneous velocity of the skeleton. Our edge cost is written as:
\begin{eqnarray}\label{eq:Cut_Edge}
E_e\big(c_i(f),c_j(f+1)\big) = \lambda_3 \bigg|\frac{K(f+1)-K(f)}{\Delta(f+1,f)}\bigg|^2 \\+ \lambda_4 \gamma  [c_i(f)\neq c_j(f+1)],\nonumber
\end{eqnarray}
where $[.]$ is the Iverson bracket, and $(\lambda_3, \lambda_4)$ are the weights between two error terms. 

We compute the smallest path using Dijkstra's shortest path algorithm. Fig.~\ref{fig:DirectCut_Snow} shows a camera switching case where it correctly switches the camera upon inter-human occlusion. Since this algorithm does not plans several steps ahead, we observe that abrupt camera switching still occurs. We smooth the path using an average filter in the post-processing stage.  

\section{Discussion and Conclusion}
\noindent We have presented a simple but powerful framework for scene-adaptive person descriptor. This is demonstrated in challenging scenes captured by mobile cameras. The learned descriptor reliably associates the same person over distant space and time instances. Our descriptor outperforms the baseline by $18$\% and our 3D skeleton reconstruction is 5-10X more stable than naive reconstruction even with ground truth people correspondences on events captured in the wild. Our algorithm works even with few cameras. This enables has potential applications to broadcast sport (e.g., basketball or football) with domain adaptation using prior \emph{unlabeled} footage as fine-tuning on a small subset of the test sequence suffices for generalization.  

The main limitation of our framework is the need for accurate detection of semantic keypoints. Unless these keypoints are well localized, our multiview synchronization and triplet mining will break down. Tracklet generation is also crucial for descriptor bootstrapping. Noisy tracklets can severely degrade the descriptor discrimination. While more sophisticated algorithms could be used to improve the tracklet generation quality~\cite{dehghan2015gmmcp,dehghan2015target}, the problem may still remain for scenes with people wearing similar and textureless clothing. One prominent solution is the use of robust estimator for the distance metric loss under the graduated non-convexity framework~\cite{shah2017robust,barron2019general}.

\section*{Acknowledgement}
This research is supported by NSF CNS-1446601, ONR N00014-14-1-0595, Heinz Endowments ``Platform Pittsburgh'', Metro 21 grants, and an Adobe Research Gift. Minh Vo was partly supported by the 2017 Qualcomm Innovation Fellowship.
\vspace{-.5cm}
{\small
\bibliographystyle{ieee}
\bibliography{egbib}
}
\boldstart{Minh Vo} is a Research Scientist at Facebook Reality Lab. He is interested in developing large-scale dynamic human scene understanding systems in order to create virtual environments that are perceptually indistinguishable from reality. He is the recipient of the 2017 Qualcomm Innovation Fellowship. He received his Ph.D. from The Robotics Institute at Carnegie Mellon University in 2019.\\

\vspace{-.1cm}
\boldstart{Ersin Yumer} is Staff Research Scientist at Uber ATG, where he leads the research team in San Francisco office. His research interests lie at the intersection of machine learning, and 3D computer vision. He develops end-to-end learning systems and holistic machine learning applications that bring signals of the visual world together: images, depth scans, videos, 3D shapes and points clouds. He has a Ph.D. from Carnegie Mellon University. \\

\vspace{-.1cm}
\boldstart{Kalyan Sunkavalli} is a Senior Research Scientist at Adobe Research. His research interests lie at the intersection of computer vision, graphics, and machine learning and focus on understanding, reconstructing, and editing visual appearance from images and videos. He received his Ph.D. from Harvard University in 2012 and Masters in 2006 from Columbia University.\\

\vspace{-.1cm}
\boldstart{Sunil Hadap} is a Principal Applied Scientist at Amazon Lab126. His research interests are in 3D object compositing, image decomposition, and depth-based image editing. His research interests include Computational Imaging/Photography, Simulation based Design Tools, and 3D Acquisition. He received the Ph.D. degree from MIRALab, the University of Geneva in 2003.\\

\vspace{-.1cm}
\boldstart{Yaser Sheikh} is the Director of the Facebook Reality Lab in Pittsburgh and is an Associate Professor at the Robotics Institute, Carnegie Mellon University. His research is broadly focused on machine perception of social behavior, spanning computer vision, computer graphics, and machine learning. His works have won Popular Science’s “Best of What’s New” Award, the Honda Initiation Award (2010), Best Student Paper Award (CVPR 2018), Best Paper Awards (WACV 2012), SAP (2012), SCA (2010), ICCV THEMIS (2009), Best Demo Award (ECCV 2016), and placed first in the MSCOCO Keypoint Challenge (2016), the Hillman Fellowship (2004). His research has been covered in popular press including NY Times, The Verge, Popular Science, BBC, MSNBC, New Scientist, Slashdot, and WIRED. He obtained his Ph.D. from University of Central Florida in 2006.\\

\vspace{-.1cm}
\boldstart{Srinivasa G. Narasimhan} is a Professor of Robotics at Carnegie Mellon University. His group focuses on imaging, illumination and light transport to enable applications in vision, graphics, robotics, agriculture, and medical imaging. His works have won Best Paper Award (CVPR 2019), Best Demo Award (ICCP 2015), A9 Best Demo Award (CVPR 2015), Marr Prize Honorable Mention Award (2013), FORD URP Award (2013), Best Paper Runner up Prize (ACM I3D 2013), Best Paper Honorable Mention Award (ICCP 2012), Best Paper Award (IEEE PROCAMS 2009), the Okawa Research Grant (2009), the NSF CAREER Award (2007), Adobe Best Paper Award (ICCVW 2007) and Best Paper Honorable Mention Award (CVPR 2000). He obtained his Ph.D. from Columbia University in 2003.\vspace{-.1cm}\vspace{-.1cm}\\
\end{document}